\documentclass{article}

 \usepackage[eandd, final]{neurips_2026}

\usepackage{titletoc}
\usepackage{wrapfig}
\usepackage[utf8]{inputenc} 
\usepackage[T1]{fontenc}    
\usepackage{hyperref}       
\usepackage{url}            
\usepackage{booktabs}       
\usepackage{amsfonts}       
\usepackage{nicefrac}       
\usepackage{microtype}      
\usepackage{xcolor}         
\usepackage[most]{tcolorbox}

\usepackage{pifont}
\newcommand{\cmark}{\ding{51}}
\newcommand{\xmark}{\ding{55}}
\usepackage{longtable}
\usepackage{pdflscape}
\usepackage{tcolorbox}
\usepackage{graphicx}
\usepackage{amsmath}
\usepackage{enumitem}
\usepackage{multirow}
\usepackage[normalem]{ulem}

\usepackage{titlesec}

\titlespacing{\section}{0pt}{1.2ex plus .2ex minus .2ex}{0.6ex}
\titlespacing{\subsection}{0pt}{1.0ex plus .2ex minus .2ex}{0.4ex}
\titlespacing{\subsubsection}{0pt}{0.8ex plus .2ex minus .2ex}{0.3ex}

\usepackage{enumitem}
\setlist{nosep}

\renewcommand{\arraystretch}{1.0}

\usepackage[font=small,skip=2pt]{caption}
\title{PlantMarkerBench: A Multi-Species Benchmark for Evidence-Grounded Plant Marker Reasoning}

\author{
Sajib Acharjee Dip$^{1}$ \\
\texttt{sajibacharjeedip@vt.edu}
\And
Song Li$^{2}$\thanks{Corresponding author.} \\
\texttt{songli@vt.edu}
\And
Liqing Zhang$^{1,3,4,5}$\footnotemark[1] \\
\texttt{lqzhang@cs.vt.edu}
\\[0.8em]
$^{1}$Department of Computer Science, Virginia Tech \\
$^{2}$School of Plant and Environmental Sciences, Virginia Tech \\
$^{3}$Health Sciences, Virginia Tech \\
$^{4}$Fralin Biomedical Research Institute, Virginia Tech \\
$^{5}$FBRI Cancer Research Center, Washington, DC\\[0.8em]
\small Dataset:\textcolor{red}{ \url{https://huggingface.co/datasets/Sajib-006/PlantMarkerBench}
}
}

\begin{document}

\maketitle

\begin{abstract}
Cell-type-specific marker genes are fundamental to plant biology, yet existing resources primarily rely on curated databases or high-throughput studies without explicitly modeling the supporting evidence found in scientific literature. We introduce PlantMarkerBench, a multi-species benchmark for evaluating literature-grounded plant marker evidence interpretation from full-text biological papers. PlantMarkerBench is constructed using a modular curation pipeline integrating large-scale literature retrieval, hybrid search, species-aware biological grounding, structured evidence extraction, and targeted human review. The benchmark spans four plant species---Arabidopsis, maize, rice, and tomato---and contains 5,550 sentence-level evidence instances annotated for marker-evidence validity, evidence type, and support strength. We define two benchmark tasks: determining whether a candidate sentence provides valid marker evidence for a gene--cell-type pair, and classifying the evidence into expression, localization, function, indirect, or negative categories. We benchmark diverse open-weight and closed-source language models across species and prompting strategies. Although frontier models achieve relatively strong performance on direct expression evidence, performance drops substantially on functional, indirect, and weak-support evidence, with evidence-type confusion emerging as a dominant failure mode. Open-weight models additionally exhibit elevated false-positive rates under ambiguous biological contexts. PlantMarkerBench provides a challenging and reproducible evaluation framework for literature-grounded biological evidence attribution and supports future research on trustworthy scientific information extraction and AI-assisted plant biology.
\end{abstract}

\section{Introduction}

Cell-type marker genes are central to plant biology, enabling the identification and characterization of cellular states across tissues, developmental stages, and environmental conditions~\cite{denyer2019spatiotemporal, jean2019dynamics, shulse2019high}. Marker genes play a key role in plant single-cell transcriptomics, spatial biology, developmental genetics, and comparative cell atlas construction~\cite{richard2016single, ryu2019single, jin2022pcmdb}. As plant single-cell datasets rapidly expand across species and modalities~\cite{chen2021plantscrnadb, he2024sty, rhee2019towards}, reliable marker identification has become increasingly important for cell-type annotation and downstream biological interpretation~\cite{stuart2019comprehensive, hao2021integrated}.

Despite the growing number of plant marker databases and atlases~\cite{jin2022pcmdb, chen2021plantscrnadb, he2024sty}, identifying reliable markers from literature remains difficult. Marker evidence is often heterogeneous and distributed across expression analysis, localization experiments, mutant phenotypes, developmental studies, and indirect biological observations~\cite{brady2007high, birnbaum2003gene, cartwright2009reconstructing}. Importantly, co-occurrence of a gene and a cell type does not necessarily imply valid marker evidence. Correct interpretation frequently requires contextual biological inference, including distinguishing direct from indirect evidence, resolving species and gene-alias ambiguity, interpreting perturbation studies, and rejecting unsupported or noisy statements~\cite{bretonnel2014biomedical, huang2023towards, guu2020retrieval}.

Recent advances in large language models (LLMs) have created new opportunities for automated biological literature understanding~\cite{achiam2023gpt, touvron2023llama, hui2024qwen2, guo2025deepseek}. However, existing evaluations in plant biology largely focus on entity extraction, marker lookup, or expression-based annotation~\cite{jin2022pcmdb, he2024sty}. Current resources do not evaluate whether a model can correctly interpret literature evidence, determine whether it supports a gene--cell-type association, classify the evidence type, and reject biologically misleading claims. As a result, the ability of modern language models to perform reliable literature-grounded biological evidence attribution remains unclear.

To address this gap, we introduce \textbf{PlantMarkerBench}, a multi-species benchmark for literature-grounded plant marker evidence attribution from full-text scientific papers. PlantMarkerBench spans four plant species---\textit{Arabidopsis thaliana}, maize, rice, and tomato---and contains 5,550 sentence-level evidence instances covering 1,036 unique genes and 127 observed cell types. Each instance is annotated for marker-evidence validity, evidence type, and support strength across biologically meaningful categories including expression, localization, functional, indirect, and negative evidence.

We construct PlantMarkerBench using a reproducible modular curation pipeline integrating full-text retrieval, species-aware biological grounding, hybrid retrieval, structured evidence grading, aggregation, and targeted human review~\cite{lewis2020retrieval, izacard2022few, yao2022react}. In the current benchmark release, we formally evaluate two core tasks: (1) marker-evidence validity prediction and (2) evidence-type classification. The released pipeline additionally supports extensible downstream curation tasks including evidence aggregation and marker verification.

\begin{figure*}[t]
    \centering
    \includegraphics[width=\textwidth]{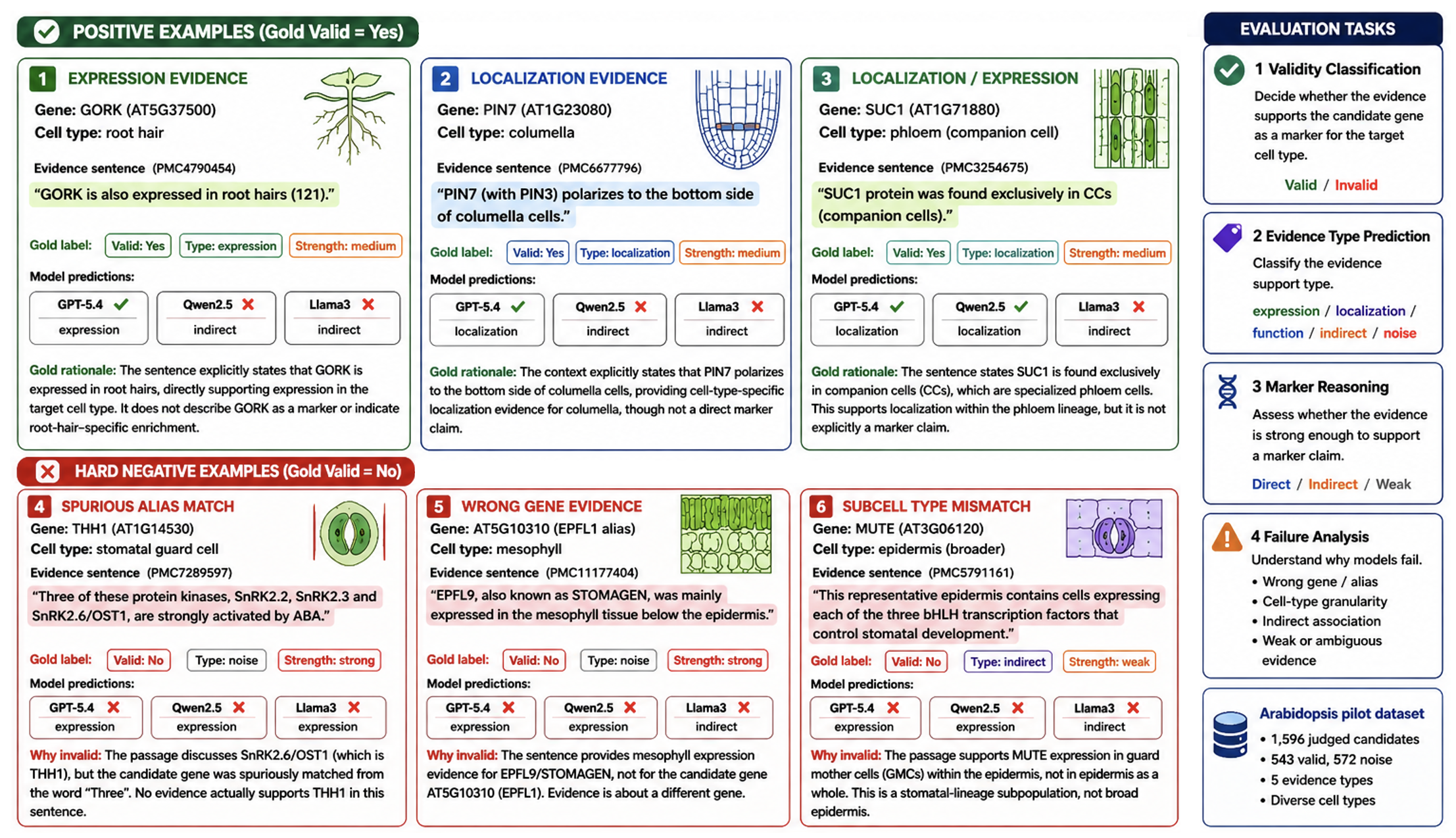}
    \caption{
\textbf{Example evidence-grounded reasoning instances in PlantMarkerBench.}
Positive examples include expression and localization evidence supporting gene--cell-type associations. Hard negative examples illustrate biologically challenging failure modes including spurious alias matching, wrong-gene attribution, and cell-type granularity mismatch. PlantMarkerBench evaluates whether models can ground the correct gene and cell type, classify evidence type, and reject misleading biological context.
}
    \label{fig:case_study}
\end{figure*}

Using PlantMarkerBench, we systematically evaluate both closed-source and open-weight LLMs across species and prompting strategies. Our experiments show that the benchmark remains challenging even for frontier models. Although strong models achieve relatively good performance on direct expression evidence, performance drops substantially on functional, indirect, and weak-support evidence, with evidence-type confusion emerging as a dominant failure mode. Figure~\ref{fig:case_study} shows representative examples from the benchmark, including biologically challenging hard negatives involving spurious aliases, wrong-gene evidence, and cell-type ambiguity.

Our contributions are summarized as follows:
\begin{itemize}
    \item We introduce PlantMarkerBench, to our knowledge, the first multi-species benchmark for literature-grounded plant marker evidence attribution from full-text scientific literature.
    
    \item We develop a reproducible modular curation pipeline integrating biological grounding, hybrid retrieval, structured evidence grading, aggregation, and targeted human review.
    
    \item We define biologically meaningful evidence regimes spanning expression, localization, functional, indirect, and negative evidence for fine-grained evaluation beyond entity extraction.
    
    \item We benchmark closed-source and open-weight LLMs across multiple prompting strategies and analyze biological failure modes through evidence-type and error-taxonomy evaluation.
\end{itemize}

\begin{table*}[ht]
\centering
\small
\setlength{\tabcolsep}{6pt}
\renewcommand{\arraystretch}{1.18}
\caption{
Comparison with related plant marker and single-cell resources.
Existing resources mainly support marker lookup, expression visualization, or atlas exploration, whereas PlantMarkerBench targets sentence-level evidence reasoning and LLM benchmarking.
}
\label{tab:related_dataset_comparison}
\resizebox{\textwidth}{!}{
\begin{tabular}{lccccc}
\toprule
\textbf{Resource} &
\textbf{Domain} &
\textbf{Species / Scale} &
\textbf{Marker Catalog} &
\textbf{Evidence Labels} &
\textbf{Primary Focus} \\
\midrule

CellMarker 2.0 &
Human / mouse &
Human, mouse; 36.3K entries &
\cmark &
\xmark &
Marker database \\

PCMDB / PlantCellMarker &
Plant markers &
6 plants; 81.1K genes; 263 cell types &
\cmark &
\xmark &
Plant marker catalog \\

PlantscRNAdb &
Plant scRNA-seq &
4 plants; 26.3K genes; 128 cell types &
\cmark &
\xmark &
scRNA marker atlas \\

PlantscRNAdb 4.0 &
Plant scRNA-seq &
33 plants; 107 datasets &
\cmark &
\xmark &
scRNA database \\

scPlantDB &
Plant single-cell atlas &
17 plants; $\sim$2.5M cells &
\cmark &
\xmark &
Single-cell atlas \\

Plant Cell Atlas resources &
Plant single-cell resources &
Multiple tools / repositories &
\cmark &
-- &
Community resource \\

\midrule

\textbf{PlantMarkerBench} &
\textbf{Plant literature} &
\textbf{4 plants; 5,550 instances; 1,036 genes} &
\cmark &
\cmark &
\textbf{Evidence benchmark} \\

\bottomrule
\end{tabular}
}
\end{table*}

\begin{table*}[ht]
\centering
\small
\setlength{\tabcolsep}{5pt}
\renewcommand{\arraystretch}{1.12}
\caption{
\textbf{PlantMarkerBench dataset statistics across four plant species.}
The benchmark contains sentence-level literature evidence annotated for marker validity, evidence type, and support strength.
Unlike binary retrieval datasets, PlantMarkerBench includes diverse biological evidence regimes together with substantial hard-negative and weak-support examples derived from full-text scientific literature.
}
\label{tab:dataset_stats}
\resizebox{\textwidth}{!}{
\begin{tabular}{lrrrrrrrr}
\toprule
\textbf{Species} &
\textbf{Full} &
\textbf{Pilot} &
\textbf{Valid} &
\textbf{Invalid} &
\textbf{Genes} &
\textbf{Cell Types} &
\textbf{Support Strength} &
\textbf{Evidence Distribution} \\
\midrule

Arabidopsis &
1,596 &
600 &
543 &
1,053 &
361 &
24 / 36 &
S:229, M:418, W:949 &
I:453, F:333, N:572, E:139, L:97 \\

Maize &
1,027 &
600 &
341 &
686 &
257 &
34 / 46 &
S:95, M:285, W:647 &
I:335, N:319, F:259, E:101, L:13 \\

Rice &
1,974 &
600 &
634 &
1,340 &
291 &
34 / 38 &
S:180, M:494, W:1300 &
N:778, I:504, F:336, E:271, L:85 \\

Tomato &
953 &
600 &
310 &
643 &
127 &
35 / 49 &
S:79, M:211, W:663 &
I:307, N:306, F:208, E:128, L:4 \\

\midrule

Total &
5,550 &
2,400 &
1,828 &
3,722 &
1,036 &
127 / 169 &
S:583, M:1,408, W:3,559 &
-- \\

\bottomrule
\end{tabular}
}
\vspace{0.5em}

\footnotesize{
E: expression, L: localization, F: function, I: indirect, N: noise.
Cell types are reported as observed benchmark cell types / total curated species vocabulary.
Support strength denotes strong (S), medium (M), and weak (W) evidence annotations.
}
\end{table*}

\begin{figure*}[t]
\centering
\includegraphics[width=\textwidth]{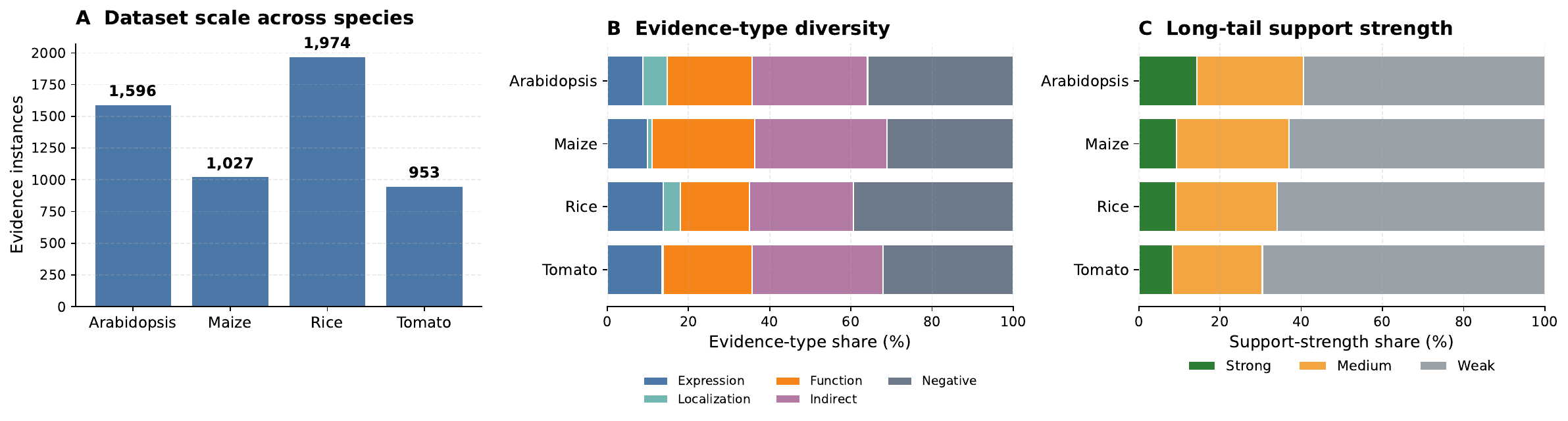}
\caption{
\textbf{PlantMarkerBench dataset overview.}
(A) Dataset scale across four plant species.
(B) Evidence-type composition showing diverse biological reasoning regimes including expression, localization, functional, indirect, and negative evidence.
(C) Long-tail support-strength distributions reveal that most literature evidence is weakly supported, reflecting realistic scientific ambiguity.
}
\label{fig:dataset_overview}
\end{figure*}

\section{Dataset Overview}

PlantMarkerBench is a multi-species benchmark for literature-grounded plant marker evidence attribution. Given a gene, candidate cell type, and evidence window, a model must determine whether the text supports the gene as a valid marker and classify the evidence type. Table~\ref{tab:dataset_stats} summarizes the release: 5,550 sentence-level evidence instances across \textit{Arabidopsis thaliana}, maize, rice, and tomato, covering 1,036 unique genes and 127 observed cell types mapped to 169 curated species-specific cell-type concepts. For controlled LLM evaluation, we construct balanced pilot subsets with 2,400 manually reviewed instances.

Unlike marker resources focused mainly on positive associations, PlantMarkerBench explicitly includes realistic literature noise, weak grounding, indirect associations, and hard negatives. Roughly two-thirds of instances are invalid, weak, indirect, or ambiguous, reflecting the difficulty of extracting reliable marker evidence from scientific papers. The dataset also spans diverse evidence regimes, including expression, localization, functional, indirect, and negative evidence. Its long-tail structure makes the benchmark especially challenging: weak-support evidence dominates, localization evidence is sparse, and indirect/functional cases require contextual biological interpretation beyond gene--cell-type co-occurrence.

\textbf{Agentic curation pipeline.}
We use a modular agentic pipeline in which specialized components exchange structured intermediate artifacts. A retrieval agent identifies candidate evidence windows, a grounding agent maps species-specific genes and cell types, an evidence-grading agent assigns structured labels and rationales, and an aggregation agent consolidates evidence across papers into marker candidates and evidence graphs. The pipeline proceeds through five stages: full-text literature filtering, species assignment, biological grounding, hybrid retrieval and candidate generation, and evidence grading with human quality control. Each stage saves auditable outputs, enabling reproducibility, targeted review, and future replacement of individual components.

\begin{figure*}[t]
\centering
\includegraphics[width=\textwidth]{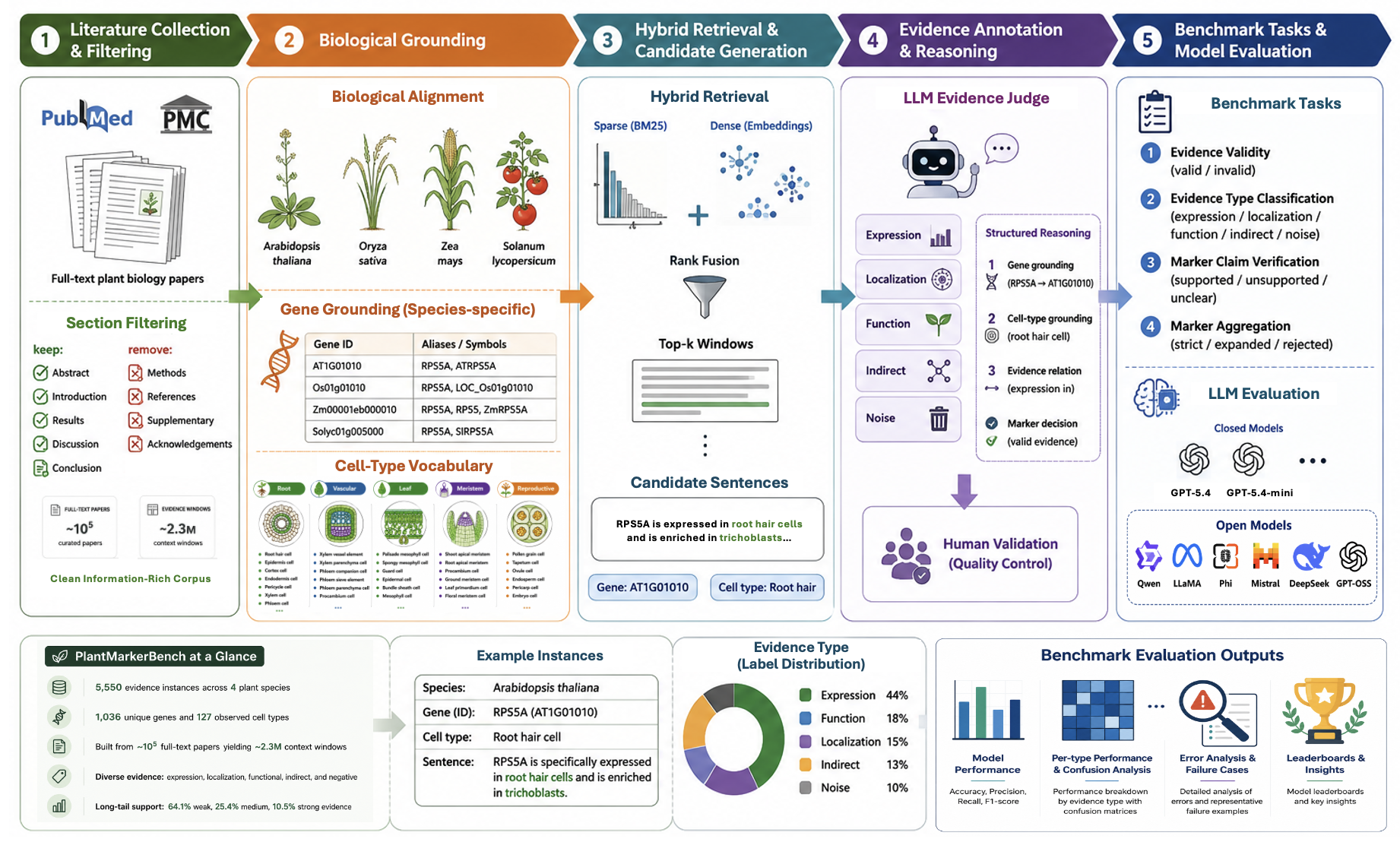}
\caption{
\textbf{PlantMarkerBench dataset overview and benchmark composition.}
PlantMarkerBench is a multi-species, evidence-grounded benchmark for plant cell-type marker reasoning constructed from full-text literature across four plant species: \textit{Arabidopsis thaliana}, maize, rice, and tomato. 
The benchmark contains 5,550 sentence-level evidence instances spanning 1,036 unique genes and 127 observed cell types. 
Evidence instances are categorized into five biologically motivated evidence regimes: expression, localization, functional, indirect, and negative/noise evidence, with accompanying support-strength annotations (strong, medium, weak). 
The benchmark is derived from approximately $10^5$ curated full-text papers and $\sim$2.3M retrieved context windows, enabling evaluation of evidence grounding, biological reasoning, and robustness under realistic literature noise and ambiguity.
}
\label{fig:pipeline}
\end{figure*}

\section{PlantMarkerBench Construction and Task Formulation}

Figure~\ref{fig:pipeline} summarizes the scale, evidence diversity, and literature-grounding characteristics of PlantMarkerBench across all four species.

\subsection{Literature Collection and Full-Text Filtering}

We collect candidate papers from PubMed and PMC using species- and cell-type-oriented queries. For papers with PMC identifiers, we download full-text XML together with metadata including title, journal, DOI, PMID, and PMCID.

To reduce irrelevant text, we retain sections likely to contain biological evidence, including abstracts, introductions, results, discussions, and conclusions, while excluding methods, references, acknowledgments, and supplementary material. We additionally filter papers with insufficient full-text content using paragraph and character-count thresholds, producing a cleaned corpus for downstream retrieval.

\subsection{Species Assignment}

Because many papers mention multiple plant species, we assign each article to a primary species before gene grounding. Species scores are computed from title, abstract, and early full-text mentions, with higher weight assigned to title and abstract occurrences. Articles without a reliable species signal are excluded to reduce cross-species contamination.

\subsection{Biological Grounding}

\subsubsection{Species-Specific Gene Matching}

Plant gene names are highly species-specific, ambiguous, and inconsistently represented across literature and databases~\citep{berardini2015arabidopsis,kawahara2013improvement,portwood2019maizegdb,fernandez2015sol}. We therefore construct a separate gene matcher for each species, mapping canonical identifiers to symbols and aliases observed in annotation resources and literature.

For Arabidopsis, we use TAIR AGI identifiers and curated symbols from TAIR~\citep{berardini2015arabidopsis}. For rice, we integrate RAP, MSU/LOC, and IC4R mappings~\citep{kawahara2013improvement,ic4r2016information}. For maize, we combine B73 v5 identifiers with curated aliases from MaizeGDB~\citep{portwood2019maizegdb}. For tomato, we integrate Solyc identifiers with SGN annotations and a conservative literature-derived lexicon~\citep{fernandez2015sol}. Each matcher stores: \texttt{gene\_id, symbol, match\_aliases}. During candidate generation, aliases are matched against evidence windows to ground mentions to species-specific canonical identifiers.

\subsubsection{Cell-Type Vocabulary Construction}

We define species-specific controlled vocabularies using terminology from plant developmental biology literature, single-cell atlases, and curated marker resources~\citep{denyer2019spatiotemporal,zhang2019single,chen2021plantscrnadb}. The vocabularies include root, vascular, leaf, meristematic, reproductive, and species-specific tissue cell types, and are used for both retrieval and gene--cell-type grounding.

\subsection{Hybrid Retrieval and Candidate Generation}
\label{sec:hybrid_retrieval}

Given a species-specific corpus and cell-type vocabulary, the retrieval agent first decomposes each article into sentence-centered evidence windows. Each window contains a target sentence and local context from adjacent sentences. Windows are filtered using noise rules that remove references, boilerplate metadata, method-heavy fragments, figure-only text, and citation-like passages. We run the same retrieval script for each species with species-specific parsed PMC files, gene matcher TSVs, and cell-type vocabularies. The pipeline outputs windows, retrieval files, broad candidates, judged evidence, and marker aggregation files.

We score evidence windows using four complementary retrieval strategies: keyword matching, BM25 sparse retrieval \citep{robertson2009probabilistic}, dense embedding retrieval \citep{reimers2019sentence}, and hybrid fusion. Keyword retrieval prioritizes co-occurrence of cell-type terms and marker-related evidence cues such as \textit{marker}, \textit{specifically expressed}, \textit{localized to}, \textit{required for}, \textit{promoter activity}, and \textit{mutant}. BM25 captures exact lexical overlap with gene and cell-type queries, while dense retrieval captures semantically related evidence. Hybrid retrieval combines sparse, dense, keyword, cell-type, and evidence-cue scores, weighted by section reliability.

For each retrieved window, the grounding agent identifies gene mentions using the species-specific gene matcher and cell-type mentions using the controlled vocabulary. Candidate instances are generated for grounded gene--cell-type pairs and deduplicated by paper, window, gene, and cell type. Each candidate retains retrieval provenance, including retrieval mode, retrieval score, section, matched alias, target sentence, and local context.

\subsection{Evidence Labeling and Aggregation}
\label{sec:evidence_labeling}

Each candidate instance is evaluated by an LLM-based grading agent using the target sentence, local evidence window, grounded gene identifier, cell type, and retrieval metadata. The grader outputs a structured JSON record containing evidence validity, evidence type, support strength, confidence, and a short rationale.

We define five evidence categories: \textbf{expression}, \textbf{localization}, \textbf{function}, \textbf{indirect}, and \textbf{negative/noise}. Direct marker mentions are normalized into the expression category. The grader is instructed to be conservative: simple gene--cell-type co-occurrence, homology-only statements, and generic developmental evidence without cell-type specificity are not treated as direct marker evidence.

To support downstream curation, judged evidence is aggregated by gene--cell-type pair into evidence graphs linking genes, evidence instances, papers, and cell types. The aggregation stage produces strict markers, expanded candidate associations, functional regulators, and indirect biological associations together with supporting evidence, provenance, and confidence statistics.

\subsection{Human Review Protocol}
\label{app:human_review}

Human quality control was performed by two reviewers with computational biology and plant single-cell analysis experience. Review focused on difficult or high-risk cases, including spurious aliases, wrong-gene grounding, cross-species ambiguity, indirect biological associations, and cell-type granularity mismatch. The pilot benchmark split was manually inspected to remove malformed or clearly unsupported instances before final release. Disagreements were resolved through discussion and adjudication using the underlying paper context and supporting evidence windows. The final benchmark therefore combines automated large-scale evidence extraction with targeted expert verification for difficult biological reasoning cases.

\subsection{Structured Reasoning Annotation}

Each instance additionally contains a structured reasoning trace decomposing the decision into four steps: gene grounding, cell-type grounding, evidence classification, and final marker decision. This provides explicit, machine-readable reasoning structure without relying on free-form chain-of-thought annotations. The pipeline is intentionally artifact-rich: intermediate outputs from retrieval, grounding, grading, and aggregation are preserved to support auditing, rerunning, and targeted correction of noisy literature-derived evidence.

\subsection{Benchmark Tasks and Evaluation Splits}
PlantMarkerBench currently supports two primary benchmark tasks:

\begin{enumerate}
    \item \textbf{Marker-evidence validity prediction}: determine whether a candidate sentence provides valid evidence supporting a gene as a marker for a target cell type.
    
    \item \textbf{Evidence-type classification}: classify the evidence into expression, localization, function, indirect, or noise categories.
\end{enumerate}

In addition, the released pipeline supports extensible downstream tasks including evidence aggregation, marker ranking, and literature-assisted curation, which are not formally benchmarked in the current release. For efficient and controlled model evaluation, we construct a balanced pilot split for Arabidopsis containing 600 examples, with equal numbers of valid and invalid evidence instances. This balanced setting enables stable comparison of precision, recall, and F1 across models. We also retain the full automatically labeled evidence set to support future evaluation under the natural class distribution. The same construction procedure is applied to rice, maize, and tomato to produce multi-species benchmark splits.

\section{Benchmarking Results}

\begin{table*}[t]
\centering
\footnotesize
\setlength{\tabcolsep}{4pt}
\renewcommand{\arraystretch}{1.08}
\caption{
\textbf{Main PlantMarkerBench leaderboard on Arabidopsis and maize.}
Open-weight models are evaluated with the default prompt and closed OpenAI models with the direct prompt.
Best scores within each species are bolded and second-best scores are underlined.
}
\label{tab:main_leaderboard_default}
\resizebox{\textwidth}{!}{
\begin{tabular}{ll l rrrrrrr}
\toprule
Species & Type & Model & Valid F1 & Evid. Macro-F1 & Expression & Localization & Function & Indirect & Negative \\
\midrule
\multirow{17}{*}{Arabidopsis}
& \multirow{2}{*}{Closed}
& GPT-5.4 & 0.549 & \textbf{0.639} & \textbf{0.754} & \uline{0.686} & \textbf{0.710} & 0.358 & \textbf{0.686} \\
&& GPT-5.4-mini & 0.633 & 0.515 & \uline{0.707} & 0.607 & 0.561 & 0.147 & 0.552 \\
\cmidrule(lr){2-10}
& \multirow{15}{*}{Open}
& Qwen2.5-32B-Instruct & \textbf{0.754} & \uline{0.545} & 0.634 & \textbf{0.715} & \uline{0.634} & 0.354 & 0.387 \\
&& GPT-OSS-20B & 0.673 & 0.436 & 0.578 & 0.033 & 0.615 & 0.332 & \uline{0.619} \\
&& Phi3-14B & 0.689 & 0.420 & 0.529 & 0.574 & 0.595 & 0.150 & 0.252 \\
&& Qwen2.5-14B-Instruct & 0.514 & 0.412 & 0.591 & 0.456 & 0.112 & \uline{0.363} & 0.538 \\
&& Llama3.2-3B & 0.449 & 0.317 & 0.380 & 0.491 & 0.434 & 0.282 & 0.000 \\
&& Llama3.1-8B & 0.634 & 0.290 & 0.436 & 0.216 & 0.402 & \textbf{0.365} & 0.033 \\
&& DeepSeek-R1-8B & \uline{0.735} & 0.268 & 0.426 & 0.154 & 0.291 & 0.277 & 0.191 \\
&& Qwen2.5-3B-Instruct & 0.070 & 0.241 & 0.158 & 0.000 & 0.362 & 0.186 & 0.501 \\
&& Mistral-7B-Instruct & 0.690 & 0.238 & 0.395 & 0.253 & 0.512 & 0.031 & 0.000 \\
&& Qwen2.5-7B-Instruct & 0.668 & 0.222 & 0.320 & 0.371 & 0.406 & 0.016 & 0.000 \\
&& Phi3-3.8B-Instruct & 0.662 & 0.199 & 0.355 & 0.229 & 0.392 & 0.017 & 0.000 \\
&& DeepSeek-R1-1.5B & 0.666 & 0.073 & 0.254 & 0.095 & 0.000 & 0.017 & 0.000 \\
&& Llama3.2-1B & 0.667 & 0.050 & 0.251 & 0.000 & 0.000 & 0.000 & 0.000 \\
&& Qwen2.5-0.5B & 0.667 & 0.050 & 0.251 & 0.000 & 0.000 & 0.000 & 0.000 \\
&& Qwen2.5-1.5B-Instruct & 0.667 & 0.050 & 0.251 & 0.000 & 0.000 & 0.000 & 0.000 \\
\midrule
\multirow{17}{*}{Maize}
& \multirow{2}{*}{Closed}
& GPT-5.4 & 0.409 & \textbf{0.460} & \textbf{0.682} & 0.000 & \textbf{0.716} & \uline{0.347} & \textbf{0.557} \\
&& GPT-5.4-mini & 0.451 & 0.400 & \uline{0.658} & 0.000 & \uline{0.658} & 0.170 & \uline{0.515} \\
\cmidrule(lr){2-10}
& \multirow{15}{*}{Open}
& Qwen2.5-32B-Instruct & \textbf{0.721} & \uline{0.443} & 0.538 & 0.240 & 0.648 & 0.298 & 0.488 \\
&& GPT-OSS-20B & 0.595 & 0.414 & 0.602 & 0.000 & 0.638 & 0.311 & 0.517 \\
&& Qwen2.5-3B-Instruct & 0.247 & 0.348 & 0.447 & \uline{0.300} & 0.586 & 0.012 & 0.394 \\
&& Llama3.2-3B & 0.253 & 0.346 & 0.485 & \textbf{0.333} & 0.595 & 0.316 & 0.000 \\
&& Llama3.1-8B & 0.445 & 0.285 & 0.504 & 0.143 & 0.397 & \textbf{0.379} & 0.000 \\
&& Mistral-7B-Instruct & 0.663 & 0.268 & 0.434 & 0.091 & 0.643 & 0.171 & 0.000 \\
&& DeepSeek-R1-8B & \uline{0.697} & 0.245 & 0.391 & 0.105 & 0.375 & 0.270 & 0.084 \\
&& Qwen2.5-14B-Instruct & 0.390 & 0.230 & 0.594 & 0.000 & 0.000 & 0.111 & 0.446 \\
&& Phi3-14B & 0.671 & 0.230 & 0.335 & 0.000 & 0.557 & 0.067 & 0.190 \\
&& Qwen2.5-7B-Instruct & 0.671 & 0.122 & 0.266 & 0.118 & 0.213 & 0.013 & 0.000 \\
&& DeepSeek-R1-1.5B & 0.663 & 0.087 & 0.262 & 0.100 & 0.074 & 0.000 & 0.000 \\
&& Qwen2.5-1.5B-Instruct & 0.661 & 0.071 & 0.254 & 0.000 & 0.061 & 0.028 & 0.014 \\
&& Phi3-3.8B-Instruct & 0.665 & 0.061 & 0.254 & 0.000 & 0.009 & 0.000 & 0.039 \\
&& Llama3.2-1B & 0.667 & 0.050 & 0.251 & 0.000 & 0.000 & 0.000 & 0.000 \\
&& Qwen2.5-0.5B & 0.667 & 0.050 & 0.251 & 0.000 & 0.000 & 0.000 & 0.000 \\
\bottomrule
\end{tabular}
}
\end{table*}

\subsection{Current LLMs Remain Far from Solving Marker Evidence Attribution}
\label{sec:llm_results}

We evaluate a broad collection of open and closed language models on Arabidopsis and maize, the two species for which full open-model evaluation is currently complete. Table~\ref{tab:main_leaderboard_default} reports open-weight Ollama models under the default prompt and OpenAI models under direct prompting. To assess stability, we additionally compute bootstrap confidence intervals on pilot-split validity F1 scores, with ranking trends remaining consistent across resampling (Appendix~\ref{app:bootstrap_ci}).

PlantMarkerBench remains challenging even for strong frontier models. Across both species, models often achieve moderate binary validity F1 while failing to correctly identify the underlying biological evidence type. This gap suggests that many systems recognize biologically relevant context without accurately grounding gene--cell-type relationships or distinguishing mechanistic evidence categories such as expression, localization, and functional support.

Several trends emerge. First, larger open-weight models substantially outperform smaller models, with Qwen2.5-32B-Instruct achieving the strongest validity F1 among open models on both species. However, even the strongest systems exhibit substantially lower evidence-type macro-F1, indicating that fine-grained biological evidence attribution remains unsolved.

Second, many smaller models exhibit degenerate behavior, achieving superficially reasonable validity scores while collapsing on evidence-type classification, often over-predicting positive evidence or failing entirely on localization and indirect evidence. Third, the benchmark exposes strong asymmetries across evidence categories: expression evidence is consistently easier than localization or indirect evidence, while localization reasoning remains especially difficult across maize and tomato. Overall, even the best configurations achieve only moderate evidence-type macro-F1, with localization and indirect evidence frequently remaining below 0.4 for many models. 

\begin{table*}[ht]
\centering
\caption{
\textbf{Cross-species OpenAI evaluation.}
For each species and model, we report the best prompt configuration selected by evidence-type macro-F1.
}
\label{tab:cross_species_openai}

\small
\setlength{\tabcolsep}{4.8pt}
\renewcommand{\arraystretch}{1.12}

\resizebox{0.98\textwidth}{!}{
\begin{tabular}{llcccccccc}
\toprule
\textbf{Species} &
\textbf{Model} &
\textbf{Prompt} &
\textbf{Valid F1} &
\textbf{Macro-F1} &
\textbf{Expr.} &
\textbf{Loc.} &
\textbf{Func.} &
\textbf{Indirect} &
\textbf{Negative} \\
\midrule

Arabidopsis & GPT-5.4 & Direct
& 0.549 & 0.639 & 0.754 & 0.686 & 0.710 & 0.358 & 0.686 \\

Arabidopsis & GPT-5.4-mini & Few-shot
& \textbf{0.731} & 0.568 & 0.735 & \textbf{0.703} & 0.535 & 0.254 & 0.614 \\

\midrule

Maize & GPT-5.4 & Structured
& 0.446 & 0.489 & 0.684 & 0.000 & \textbf{0.695} & \textbf{0.417} & 0.648 \\

Maize & GPT-5.4-mini & Few-shot
& \textbf{0.622} & 0.453 & 0.727 & 0.353 & 0.503 & 0.181 & 0.501 \\

\midrule

Rice & GPT-5.4 & Few-shot
& \textbf{0.766} & \textbf{0.655} & \textbf{0.777} & \textbf{0.654} & \textbf{0.730} & \textbf{0.393} & \textbf{0.723} \\

Rice & GPT-5.4-mini & Structured
& 0.680 & 0.546 & 0.670 & 0.536 & 0.615 & 0.257 & 0.652 \\

\midrule

Tomato & GPT-5.4 & Structured
& 0.573 & \textbf{0.531} & \textbf{0.824} & 0.000 & \textbf{0.684} & \textbf{0.540} & \textbf{0.605} \\

Tomato & GPT-5.4-mini & Few-shot
& \textbf{0.713} & 0.478 & 0.787 & \textbf{0.333} & 0.540 & 0.215 & 0.513 \\

\bottomrule
\end{tabular}
}

\vspace{0.3em}
\footnotesize{
Expr.: expression evidence, Loc.: localization evidence, Func.: functional evidence.
}

\end{table*}

\subsection{Cross-Species Evaluation Reveals Species-Specific Grounding Challenges}
\label{sec:cross_species}

We evaluate closed models across all four species using the full prompt suite. Table~\ref{tab:cross_species_openai} reports the best-performing prompt for each species--model pair according to evidence-type macro-F1.

Performance varies substantially across species, indicating that plant-marker evidence attribution does not transfer uniformly across biological domains. Rice achieves the strongest overall evidence macro-F1 with GPT-5.4, whereas maize and tomato remain considerably more challenging, particularly for localization and indirect evidence. In several cases, localization F1 collapses despite moderate validity performance, suggesting that models often recognize biologically relevant genes while failing to resolve precise cellular grounding. These results highlight an important contribution of PlantMarkerBench: benchmark difficulty arises not only from biological reasoning itself, but also from species-specific nomenclature, synonym ambiguity, and heterogeneous literature conventions. Strong performance on one species therefore does not reliably translate to robust cross-species evidence attribution.

\begin{table}[ht]
\centering
\caption{
\textbf{Prompt ablation averaged across four species.}
Few-shot prompting consistently improves validity prediction, while evidence-type reasoning remains challenging, particularly for localization and indirect evidence.
}
\label{tab:prompt_ablation}

\small
\setlength{\tabcolsep}{4.0pt}
\renewcommand{\arraystretch}{1.10}

\resizebox{0.98\linewidth}{!}{
\begin{tabular}{llccccccc}
\toprule
\textbf{Model} &
\textbf{Prompt} &
\textbf{Valid F1} &
\textbf{Macro-F1} &
\textbf{Expr.} &
\textbf{Loc.} &
\textbf{Func.} &
\textbf{Indirect} &
\textbf{Negative} \\
\midrule

GPT-5.4 & Direct
& 0.499 & \textbf{0.552}
& 0.765 & 0.299 & \textbf{0.705} & 0.378 & 0.610 \\

GPT-5.4 & Few-shot
& \textbf{0.736} & 0.544
& \textbf{0.781} & \textbf{0.321} & 0.684 & 0.299 & 0.636 \\

GPT-5.4 & Structured
& 0.582 & 0.529
& 0.728 & 0.129 & 0.682 & \textbf{0.432} & \textbf{0.672} \\

GPT-5.4 & Conservative
& 0.432 & 0.523
& 0.681 & 0.273 & 0.623 & 0.404 & 0.635 \\

\midrule

GPT-5.4-mini & Few-shot
& \textbf{0.689} & \textbf{0.500}
& 0.738 & \textbf{0.488} & 0.517 & 0.196 & 0.563 \\

GPT-5.4-mini & Structured
& 0.638 & 0.477
& 0.666 & 0.288 & 0.560 & \textbf{0.273} & \textbf{0.598} \\

GPT-5.4-mini & Direct
& 0.574 & 0.452
& 0.720 & 0.274 & \textbf{0.583} & 0.144 & 0.541 \\

GPT-5.4-mini & Conservative
& 0.524 & 0.435
& \textbf{0.763} & 0.396 & 0.393 & 0.087 & 0.536 \\

\bottomrule
\end{tabular}
}

\vspace{0.3em}
\footnotesize{
Expr.: expression evidence, Loc.: localization evidence, Func.: functional evidence.
}

\end{table}


\subsection{Prompting Improves Validity Prediction but Not Evidence Attribution}
\label{sec:prompt_ablation}

We compare direct, structured, conservative, and few-shot prompting averaged across all four species. Table~\ref{tab:prompt_ablation} shows that few-shot prompting substantially improves binary validity F1, particularly for GPT-5.4, but does not consistently improve fine-grained evidence attribution. Direct prompting achieves the strongest average evidence macro-F1 for GPT-5.4, while few-shot prompting performs best for GPT-5.4-mini. Across both models, localization and indirect evidence remain consistently difficult despite prompt engineering. These results suggest that the primary challenge is not simply recognizing relevant biological sentences, but correctly grounding gene--cell-type relationships and mechanistic evidence categories. Overall, prompting alone appears insufficient for robust literature-grounded biological evidence attribution.

\subsection{Evidence-Type Difficulty Analysis}
\label{sec:hard_subset}

To better understand biological failure modes beyond aggregate leaderboard scores, we evaluate models on curated evidence-specific subsets. Table~\ref{tab:hard_subset_main} shows that expression evidence is consistently easier than indirect or functional evidence across nearly all models. While many systems achieve strong performance on explicit expression cues, performance drops substantially on indirect evidence requiring contextual biological interpretation. Stronger models also fail differently across evidence types. GPT-5.4 with few-shot prompting achieves the strongest overall performance, while Qwen2.5-32B-Instruct performs comparatively better on localization and indirect evidence among open models. In contrast, several smaller models achieve high expression accuracy while collapsing on indirect or weakly grounded evidence, suggesting shortcut-style prediction behavior rather than robust biological interpretation.
\begin{table*}[ht]
\centering
\small
\setlength{\tabcolsep}{6pt}
\renewcommand{\arraystretch}{1.12}

\caption{
\textbf{Hard-subset evaluation on Arabidopsis.}
We report evidence-type accuracy on biologically difficult subsets from PlantMarkerBench.
Best values are bolded and second-best values are underlined.
}

\label{tab:hard_subset_main}

\begin{tabular*}{\textwidth}{@{\extracolsep{\fill}}llcccccc}
\toprule

&
\textbf{Model} &
\textbf{Overall} &
\textbf{Expr.} &
\textbf{Loc.} &
\textbf{Func.} &
\textbf{Indirect} &
\textbf{Negative} \\

\midrule

\multirow{6}{*}{\rotatebox[origin=c]{90}{\textbf{Closed}}}

& GPT-5.4 (Few-shot)
& \underline{0.633}
& 0.791
& 0.518
& 0.643
& 0.094
& \textbf{0.948} \\

& GPT-5.4-mini (Few-shot)
& 0.577
& 0.837
& 0.696
& 0.387
& 0.180
& 0.861 \\

& GPT-5.4 (Direct)
& \textbf{0.652}
& 0.802
& 0.625
& \underline{0.744}
& 0.291
& 0.740 \\

& GPT-5.4-mini (Structured)
& 0.552
& \underline{0.919}
& 0.429
& 0.429
& 0.256
& 0.728 \\

& GPT-5.4 (Structured)
& 0.582
& 0.884
& 0.143
& 0.560
& 0.291
& 0.792 \\

& GPT-5.4-mini (Conservative)
& 0.535
& 0.861
& 0.643
& 0.298
& 0.068
& 0.884 \\

\midrule

\multirow{6}{*}{\rotatebox[origin=c]{90}{\textbf{Open}}}

& Qwen2.5-32B-Instruct
& 0.535
& \textbf{0.919}
& \textbf{0.875}
& 0.607
& \underline{0.385}
& 0.266 \\

& GPT-OSS-20B
& 0.530
& 0.814
& 0.018
& 0.619
& 0.299
& 0.624 \\

& Phi3-14B
& 0.463
& 0.907
& \underline{0.625}
& \textbf{0.762}
& 0.094
& 0.150 \\

& DeepSeek-R1-8B
& 0.305
& \textbf{0.954}
& 0.107
& 0.220
& 0.325
& 0.116 \\

& Mistral-7B-Instruct
& 0.330
& 0.942
& 0.179
& 0.625
& 0.017
& 0.000 \\

& Llama3.1-8B
& 0.332
& 0.779
& 0.143
& 0.304
& \textbf{0.598}
& 0.017 \\

\bottomrule

\end{tabular*}

\vspace{0.3em}

\footnotesize{
Overall denotes evidence-type accuracy on the full Arabidopsis split; subset columns report evidence-type accuracy within each gold evidence category.
}

\end{table*}


\subsection{Error Taxonomy Analysis}
\label{sec:error_taxonomy}

\begin{wrapfigure}{r}{0.52\textwidth}
\vspace{-0.8em}
\centering
\includegraphics[width=0.50\textwidth]{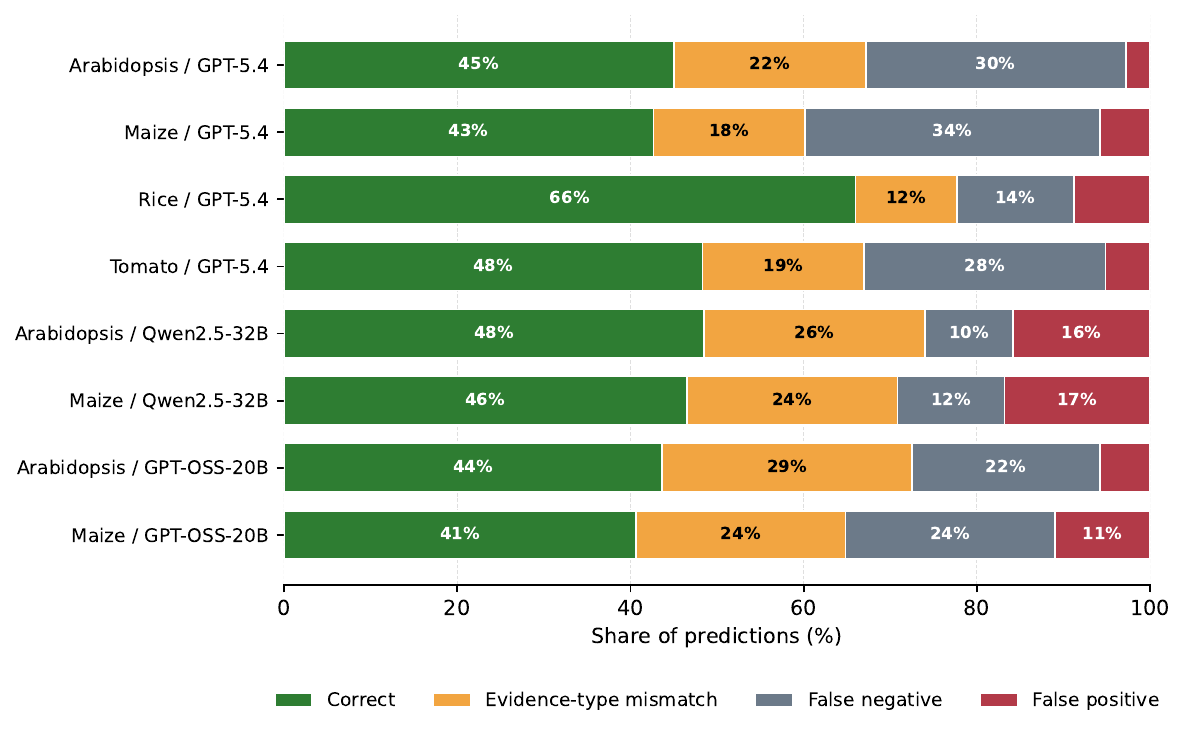}
\caption{
\textbf{Error taxonomy across representative PlantMarkerBench runs.}
Evidence-type mismatch is the dominant failure mode across most settings, while open-weight models exhibit substantially higher false-positive rates.
}
\label{fig:error_taxonomy}
\end{wrapfigure}

To better understand model behavior beyond aggregate accuracy, we analyze prediction failures across representative PlantMarkerBench runs. Figure~\ref{fig:error_taxonomy} decomposes predictions into correct predictions, evidence-type mismatches, false negatives on supported evidence, and false positives on unsupported evidence. Across nearly all settings, evidence-type mismatch is the dominant failure mode, indicating that models often recognize biologically relevant evidence but struggle to distinguish expression, localization, functional, and indirect support. Open-weight models additionally exhibit substantially higher false-positive rates, frequently over-predicting marker evidence from weak co-occurrence patterns or indirect biological associations. Smaller models further struggle with gene-alias ambiguity and fine-grained cell-type distinctions. 

\subsection{Qualitative Evidence Reasoning Analysis}
\label{sec:qualitative_analysis}

Figure~\ref{fig:case_study} shows representative reasoning instances from PlantMarkerBench. Positive examples require models to distinguish multiple evidence regimes, including expression, localization, and functional support, while correctly grounding genes and cell types from contextual biological evidence. The hard negative examples highlight common failure modes, including spurious alias matching, wrong-gene attribution within gene families, indirect associations mistaken as direct evidence, and cell-type granularity confusion. These cases demonstrate that PlantMarkerBench evaluates literature-grounded biological evidence attribution under realistic ambiguity rather than simple entity extraction or keyword matching.

\section{Conclusion}

We introduced PlantMarkerBench, a multi-species benchmark for literature-grounded plant marker evidence attribution from full-text biological literature. The benchmark spans four plant species and evaluates whether models can correctly interpret diverse evidence regimes linking genes to cell types. Our experiments show that current LLMs still struggle with fine-grained biological evidence attribution. While strong models perform reasonably well on direct expression evidence, performance drops substantially on functional, indirect, and weak-support evidence, with evidence-type confusion emerging as a dominant failure mode. Alongside the benchmark, we release a reproducible modular curation pipeline integrating retrieval, biological grounding, structured evidence grading, aggregation, and targeted human review. 

\newpage
\bibliography{main}
\bibliographystyle{plainnat}



\newpage
\appendix

\section*{Appendix}
\addcontentsline{toc}{section}{Appendix}

\startcontents[appendix]
\printcontents[appendix]{l}{1}{\setcounter{tocdepth}{2}}

\newpage
\section{Extended Dataset Construction Details}
\label{app:construction}
\begin{table}[ht]
\centering
\small
\caption{
\textbf{Pipeline outputs used for dataset construction and review.}
Each stage writes auditable artifacts used for benchmark generation, marker aggregation, and human quality control.
}
\label{tab:pipeline_outputs}
\resizebox{\linewidth}{!}{
\begin{tabular}{lll}
\toprule
Stage & Output artifact & Purpose \\
\midrule
Windowing & \texttt{windows.json} & Sentence-centered evidence windows \\
Retrieval & \texttt{retrieval\_\{mode\}.json} & Retrieved evidence windows with scores \\
Candidate generation & \texttt{broad\_candidates.json} & Gene--cell-type candidate evidence \\
Evidence grading & \texttt{judged\_evidence.json} & Structured validity/type labels \\
Evidence graph & \texttt{evidence\_graph.graphml} & Gene--evidence--cell--paper graph \\
Aggregation & \texttt{strict\_curated\_markers.json} & High-confidence expression/localization markers \\
Aggregation & \texttt{expanded\_candidate\_associations.json} & Broader valid evidence associations \\
Aggregation & \texttt{functional\_regulators.json} & Functional gene--cell-type regulators \\
Aggregation & \texttt{indirect\_associations.json} & High-confidence indirect associations \\
Reporting & \texttt{summary\_report.md} & Review-ready construction summary \\
\bottomrule
\end{tabular}
}
\end{table}

\subsection{PMC Retrieval and Literature Collection}
We collect plant biology full-text articles from the PubMed Central Open Access (PMC OA) subset using species-specific keyword queries and taxonomy-aware retrieval rules. Queries include scientific names, common names, tissue terms, developmental terminology, and marker-related biological concepts. We retain only papers with accessible full text and sufficient biological content for downstream evidence extraction.

We release the full paper lists, PMC identifiers, and retrieval scripts for reproducibility.

\subsection{Full-Text Parsing and Section Filtering}
We parse PMC XML documents and extract titles, abstracts, figure captions, and body paragraphs. To improve evidence quality, we retain sections whose titles match:
\textit{abstract},
\textit{introduction},
\textit{results},
\textit{discussion},
or
\textit{conclusion}.

We exclude sections associated with:
\textit{methods},
\textit{materials},
\textit{references},
\textit{acknowledgments},
\textit{supplementary material},
and boilerplate metadata.

Paragraphs shorter than a minimum token threshold are discarded. Papers failing minimum content-length criteria are removed from the corpus.

\subsection{Species Assignment}
Each paper is assigned to a primary species using weighted mention statistics. Mentions in titles and abstracts receive higher weight than mentions in body paragraphs. Species identification patterns include both scientific and common names, including:
\textit{Arabidopsis thaliana},
rice/\textit{Oryza sativa},
maize/\textit{Zea mays},
and tomato/\textit{Solanum lycopersicum}.

Ambiguous or mixed-species papers are filtered conservatively.

\subsection{Species-Specific Gene Matcher Construction}
We construct species-aware gene normalization resources from public annotation databases and curated synonym mappings.

\begin{itemize}
    \item \textbf{Arabidopsis}: AGI identifiers, TAIR symbols, aliases, and curated synonym mappings.
    \item \textbf{Rice}: RAP, MSU, LOC, and IC4R-aligned mappings.
    \item \textbf{Maize}: B73 v5 locus identifiers and \texttt{Zm00001eb}-style mappings aligned with common gene symbols.
    \item \textbf{Tomato}: Solyc identifiers and manually filtered tomato gene synonym lexicons.
\end{itemize}

To reduce spurious matches, we remove highly ambiguous aliases, short abbreviations, generic biological words, and cell-type-confounded aliases.

\subsection{Cell-Type Vocabulary Construction}
We manually curate plant cell-type vocabularies spanning root, vascular, leaf, reproductive, and meristematic tissues. The vocabularies include canonical names, plural forms, common abbreviations, and biologically related variants.

Representative categories include:
root hair,
endodermis,
cortex,
xylem,
phloem,
companion cell,
guard cell,
mesophyll,
columella,
pericycle,
SAM,
RAM,
pollen,
and placenta.

The full vocabularies are released with the benchmark resources.

\subsection{Hybrid Retrieval and Candidate Generation}
For each paper, we generate candidate evidence windows using a hybrid retrieval pipeline combining:

\begin{itemize}
    \item BM25 lexical retrieval,
    \item dense embedding retrieval,
    \item keyword matching,
    \item and hybrid retrieval fusion.
\end{itemize}

Candidate windows are constructed around co-occurring gene and cell-type mentions. Each candidate stores:
paper identifiers,
retrieval scores,
retrieval modes,
evidence sentences,
local context windows,
gene normalization outputs,
and matched cell-type metadata.

\subsubsection{Hybrid Retrieval Scoring}
Each evidence window receives sparse, dense, keyword, cell-type, and evidence-cue scores. Hybrid retrieval uses the following scoring function:
\[
s(w) =
\left(
0.30s_{\mathrm{BM25}}(w)
+0.30s_{\mathrm{emb}}(w)
+0.15s_{\mathrm{kw}}(w)
+0.15s_{\mathrm{cell}}(w)
+0.10s_{\mathrm{cue}}(w)
\right)
s_{\mathrm{section}}(w).
\]
Section weights prioritize results and abstracts while down-weighting introductions and methods-like passages.

\subsubsection{Candidate Generation Rules}
For each retrieved evidence window, we identify gene mentions using the species-specific matcher and cell-type mentions using the controlled vocabulary. We retain candidates with explicit gene evidence and deduplicate by paper, window, gene, and cell type. To preserve recall, up to five high-scoring genes are retained per evidence window.

\subsection{Evidence Annotation Pipeline}
Candidate evidence windows are labeled using structured LLM-based evidence grading followed by targeted human quality control and manual review.

Each candidate is annotated with:
\begin{itemize}
    \item evidence validity,
    \item evidence type,
    \item support strength,
    \item and structured rationale.
\end{itemize}

We define five evidence categories:
expression,
localization,
function,
indirect,
and noise.

Human review focuses on difficult biological edge cases including:
subcell-type mismatches,
family-level evidence,
cross-species ambiguity,
spurious alias matches,
and weak biological associations.

\subsection{Marker Aggregation Scoring}
For each gene--cell-type group, we compute a final score from evidence-type weights, average confidence, section reliability, number of supporting papers, and retrieval-mode consensus:
\[
S(g,c) =
\sum_{e \in E_{g,c}} w_{\mathrm{type}}(e)
+ b_{\mathrm{paper}}
+ b_{\mathrm{retrieval}}
+ \bar{p}_{\mathrm{conf}}
+ 0.3\bar{s}_{\mathrm{section}}.
\]
Strict markers retain only direct marker, expression, or localization evidence above threshold, whereas expanded candidates additionally include functional and high-confidence indirect evidence.

\subsection{Alias and Noise Filtering}
We remove generic aliases, short ambiguous symbols, cell-type names, section labels, citation artifacts, and common biological words that would create spurious gene matches. Sentence-level noise filters remove URLs, DOI fragments, reference-like strings, figure-only captions, journal names, correspondence metadata, and boilerplate publication text.

\subsection{Benchmark Split Construction}
We release both full evidence corpora and balanced pilot benchmark splits for controlled evaluation.

Pilot splits are constructed using balanced sampling across:
\begin{itemize}
    \item valid and invalid evidence,
    \item evidence types,
    \item species,
    \item genes,
    \item and cell types.
\end{itemize}

These splits are used for all reported benchmark evaluations.

\section{Dataset Format and Released Fields}
\label{app:dataset_schema}

PlantMarkerBench is released as sentence-level JSONL records. 
Each instance stores the grounded gene--cell-type pair, evidence sentence, local context window, structured biological labels, retrieval provenance, and optional reasoning metadata. 
The released schema is designed to support evidence reasoning, retrieval analysis, biological grounding, and reproducibility studies.

\begin{table*}[t]
\centering
\small
\setlength{\tabcolsep}{5pt}
\renewcommand{\arraystretch}{1.12}
\caption{
\textbf{Released PlantMarkerBench fields and schema.}
Each JSONL instance stores grounded biological entities, evidence context, structured labels, and provenance metadata.
}
\label{tab:dataset_schema}
\resizebox{\textwidth}{!}{
\begin{tabular}{lll}
\toprule
\textbf{Field} & \textbf{Type} & \textbf{Description} \\
\midrule
id & string & Unique benchmark instance identifier \\
species & string & Plant species name \\
paper\_id & string & Source PMC paper identifier \\
window\_id & string & Evidence-window identifier \\
candidate\_id & string & Grounded candidate identifier \\
gene\_id & string & Canonical species-specific gene identifier \\
gene\_symbol & string & Canonical gene symbol \\
matched\_alias & string & Alias matched during retrieval or grounding \\
cell\_type & string & Grounded plant cell type \\
section & string & Paper section containing the evidence window \\
target\_sentence & string & Central evidence sentence used for evaluation \\
window\_text & string & Local evidence context window \\
gold.is\_valid\_marker\_evidence & boolean & Binary marker-evidence validity label \\
gold.evidence\_type & categorical & Expression, localization, function, indirect, or noise \\
gold.support\_strength & categorical & Strong, medium, weak, or none \\
structured\_reasoning & list & Intermediate reasoning traces and grounding steps \\
rationale\_gold & string & Short biological rationale for the gold label \\
prompt & string & Evaluation prompt used during annotation or benchmarking \\
reviewer\_flag & optional boolean & Optional flag for manually reviewed difficult cases \\
retrieval\_metadata & optional dict & Retrieval scores, modes, and provenance metadata \\
\bottomrule
\end{tabular}
}
\end{table*}

\subsection{Example JSON Record}

Listing~\ref{lst:json_example} shows a representative benchmark instance.

\begin{lstlisting}[
caption={Representative PlantMarkerBench JSONL instance.},
label={lst:json_example},
basicstyle=\ttfamily\footnotesize,
breaklines=true,
columns=fullflexible
]
{
  "id": "arabidopsis_ev_000000",
  "species": "arabidopsis",
  "paper_id": "PMC3935571",
  "gene_id": "AT4G13260",
  "gene_symbol": "YUC2",
  "matched_alias": "YUC2",
  "cell_type": "pericycle",
  "section": "results",
  "target_sentence":
    "Locally induced auxin biosynthesis in a single
     pericycle cell is sufficient to initiate LRPs.",
  "gold": {
    "is_valid_marker_evidence": false,
    "evidence_type": "indirect",
    "support_strength": "weak"
  }
}
\end{lstlisting}

The released benchmark additionally includes species-level statistics, retrieval outputs, prediction files, prompt templates, evaluation scripts, and error-analysis utilities. Intermediate artifacts are preserved to support auditing and future extension of the benchmark construction pipeline.

\subsection{Evidence Label Definitions}
\label{app:labels}

\begin{table}[ht]
\centering
\small
\begin{tabular}{p{0.18\linewidth}p{0.72\linewidth}}
\toprule
\textbf{Label} & \textbf{Definition} \\
\midrule
Expression & The sentence reports gene expression, enrichment, reporter activity, or transcript/protein abundance in the target cell type. \\

Localization & The sentence reports spatial localization of a gene product, reporter signal, or protein to the target cell type. \\

Function & The sentence links the gene to development, identity, morphology, or biological function of the target cell type through perturbation or phenotype evidence. \\

Indirect & The sentence is biologically related but does not directly establish marker evidence for the target gene--cell-type pair. \\

Noise & The sentence is irrelevant, wrong-gene, wrong-species, citation-only, background-only, or otherwise unsupported. \\

\bottomrule
\end{tabular}
\caption{Evidence type definitions used in PlantMarkerBench.}
\label{tab:label_definitions}
\end{table}

\section{Dataset Statistics}
\label{app:dataset_stats}



\subsection{Cell-Type Diversity}
\label{app:cell_type_diversity}

PlantMarkerBench spans a broad range of biologically relevant plant cell types across root, vascular, epidermal, mesophyll, reproductive, and meristematic tissues. Unlike conventional marker databases that primarily focus on canonical or highly studied marker genes, PlantMarkerBench captures evidence grounded directly in full-text literature, resulting in substantial diversity in both cell-type coverage and evidence composition (Table \ref{tab:cell_type_diversity}).

The dataset additionally exhibits a pronounced long-tail distribution. While common cell types such as root hair, endodermis, cortex, xylem, guard cell, mesophyll, and phloem appear frequently, many specialized or developmentally specific cell types occur only a small number of times. This imbalance reflects realistic biological literature distributions, where experimentally tractable or historically well-studied tissues dominate published evidence. As a result, PlantMarkerBench evaluates not only performance on common marker associations, but also the ability of models to reason over sparse and heterogeneous biological evidence.

Cell-type vocabularies are constructed separately for each species and include both canonical plant anatomy terms and species-specific developmental tissues. These vocabularies are used during retrieval, grounding, candidate generation, and evaluation. Across all four species, the benchmark covers more than one hundred observed cell types with substantial variation in tissue composition and annotation density.

\begin{table}[ht]
\centering
\small
\setlength{\tabcolsep}{5pt}
\renewcommand{\arraystretch}{1.12}
\caption{
Cell-type diversity in PlantMarkerBench. Tail fraction denotes the percentage of evidence instances assigned to cell types with fewer than 10 examples.
}
\label{tab:cell_type_diversity}
\resizebox{\linewidth}{!}{
\begin{tabular}{lccp{0.58\linewidth}}
\toprule
\textbf{Species} & \textbf{Unique Cell Types} & \textbf{Tail Fraction} & \textbf{Most Frequent Cell Types} \\
\midrule
Arabidopsis & 24 & 0.9\% & root hair (262), xylem (241), stomata (170), epidermis (158), phloem (154), meristem (103), guard cell (74), cortex (67), stele (66), columella (40) \\
Maize & 34 & 5.2\% & embryo (194), stomata (113), pollen (102), endosperm (97), root hair (83), pollen tube (55), aleurone (54), mesophyll (29), xylem (27), root meristem (23) \\
Rice & 34 & 1.7\% & stomata (239), xylem (193), root hair (168), shoot apical meristem (151), epidermis (132), phloem (132), leaf sheath (123), vascular bundle (94), guard cell (68), floral meristem (64) \\
Tomato & 35 & 6.2\% & pericarp (156), trichome (140), pollen (122), anther (119), ovule (46), stomata (40), fruit pericarp (26), columella (24), pollen tube (22), tapetum (21) \\
\bottomrule
\end{tabular}
}
\end{table}



\section{Prompt Templates and Annotation Protocols}
\label{app:prompt}

\subsection{Evidence Grading Prompt}

\begin{tcolorbox}[colback=gray!5,colframe=gray!40,title=Evidence grading prompt]
You are evaluating plant cell-type marker evidence.

Given a species, gene identifier, gene symbol or alias, cell type, evidence sentence, and local context, decide:
(1) whether the sentence provides valid marker evidence for the gene--cell-type pair;
(2) the evidence type: expression, localization, function, indirect, or noise;
and (3) support strength: strong, medium, weak, or none.

Return JSON only with fields:
\texttt{is\_valid\_marker\_evidence},
\texttt{evidence\_type},
\texttt{support\_strength}, and
\texttt{rationale\_short}.
\end{tcolorbox}

\subsection{Human Review Protocol}

We provide annotation instructions and biological review criteria used during manual quality control. Reviewers evaluate:
gene specificity,
cell-type specificity,
marker strength,
cross-species ambiguity,
and evidence grounding quality.

\section{Benchmark Tasks and Evaluation Details}
\label{app:evaluation}

\subsection{Validity Classification}
Binary classification of whether a candidate evidence window supports the proposed gene--cell-type marker relationship.

\subsection{Evidence-Type Prediction}
Multi-class prediction across:
expression,
localization,
function,
indirect,
and noise evidence.

\subsection{Marker Reasoning Evaluation}
Evaluation of structured biological reasoning and rationale quality.

\subsection{Error Taxonomy}
We categorize model failures into:
false positives,
false negatives,
evidence-type mismatches,
subcell-type mismatches,
spurious alias errors,
and biologically indirect associations.


\section{Prompt Templates and Evaluation Protocols}
\label{app:prompts_protocols}

\subsection{Evaluation Prompt Modes}
\label{app:prompt_modes}

We evaluate each model using four prompt modes: direct, structured, conservative, and few-shot. 
All prompts receive the same input fields: species, gene identifier, gene symbol or alias, target cell type, target evidence sentence, and local context window. 
The model is required to return JSON only, with fields for marker-evidence validity, evidence type, support strength, and a short rationale. 
Direct marker evidence is normalized to expression evidence during evaluation.

\begin{table}[h]
\centering
\small
\setlength{\tabcolsep}{5pt}
\renewcommand{\arraystretch}{1.12}
\caption{Prompt modes used for model evaluation.}
\label{tab:prompt_modes}
\resizebox{\linewidth}{!}{
\begin{tabular}{lp{0.70\linewidth}}
\toprule
\textbf{Prompt Mode} & \textbf{Description} \\
\midrule
Direct & Asks the model to decide whether the sentence provides valid marker evidence and classify the evidence type. \\
Structured & Adds an internal reasoning checklist covering gene grounding, cell-type grounding, relation type, and final marker decision. \\
Conservative & Instructs the model to reject wrong-gene, wrong-species, homology-only, background, organ-level, and citation-only evidence. \\
Few-shot & Provides short label examples for expression, localization, function, indirect, and noise evidence before evaluation. \\
\bottomrule
\end{tabular}
}
\end{table}

\subsection{Prompt Templates}
\label{app:prompts}

We evaluate multiple prompting strategies for evidence grading, ranging from direct classification to structured reasoning and conservative biological filtering. All prompts return a structured JSON prediction containing validity, evidence type, support strength, and a short rationale.

\subsubsection{Direct Prompt}

\begin{tcolorbox}[
    colback=gray!5,
    colframe=gray!40,
    title=Direct prompt,
    breakable,
    width=\linewidth
]
\footnotesize
\ttfamily

You are evaluating plant cell-type marker evidence.

Species: <species> \\
Gene ID: <gene\_id> \\
Gene symbol or alias: <gene\_symbol> \\
Cell type: <cell\_type>

\vspace{0.3em}

Evidence sentence: \\
<target\_sentence>

\vspace{0.3em}

Local context: \\
<window\_text>

\vspace{0.3em}

Decide whether this is valid marker evidence.

Evidence type must be one of:
expression, localization, function, indirect, noise.

Treat direct marker evidence as expression.

Support strength must be one of:
strong, medium, weak, none.

\vspace{0.3em}

Return JSON only:

\{
"is\_valid\_marker\_evidence": true, \\
"evidence\_type": "expression", \\
"support\_strength": "medium", \\
"rationale\_short": "one short explanation"
\}

\end{tcolorbox}

\subsubsection{Structured Reasoning Prompt}

\begin{tcolorbox}[
    colback=gray!5,
    colframe=gray!40,
    title=Structured reasoning prompt,
    breakable,
    width=\linewidth
]
\footnotesize
\ttfamily

You are evaluating plant cell-type marker evidence.

Use this structured reasoning internally:

1. Gene grounding:
does the evidence refer to the target gene or alias?

2. Cell-type grounding:
is the target cell type explicitly or experimentally connected?

3. Relation type:
expression, localization, function, indirect, or noise.

4. Final marker decision:
valid or invalid.

\vspace{0.3em}

Species: <species> \\
Gene ID: <gene\_id> \\
Gene symbol or alias: <gene\_symbol> \\
Cell type: <cell\_type>

\vspace{0.3em}

Evidence sentence: \\
<target\_sentence>

\vspace{0.3em}

Local context: \\
<window\_text>

\vspace{0.3em}

Be conservative, but do not reject valid biological evidence only
because the sentence does not explicitly use the word ``marker''.

\vspace{0.3em}

Return JSON only:

\{
"is\_valid\_marker\_evidence": true, \\
"evidence\_type": "expression", \\
"support\_strength": "medium", \\
"rationale\_short": "one short explanation"
\}

\end{tcolorbox}

\subsubsection{Conservative Prompt}

\begin{tcolorbox}[
    colback=gray!5,
    colframe=gray!40,
    title=Conservative prompt,
    breakable,
    width=\linewidth
]
\footnotesize
\ttfamily

You are evaluating plant cell-type marker evidence.

Be very conservative. Mark valid only if the evidence clearly links
the target gene to the target cell type.

Reject:
- wrong-gene evidence \\
- wrong-species evidence \\
- homology-only evidence \\
- pathway/background mentions \\
- citation/author-name noise \\
- organ-level evidence without cell-type grounding

\vspace{0.3em}

Species: <species> \\
Gene ID: <gene\_id> \\
Gene symbol or alias: <gene\_symbol> \\
Cell type: <cell\_type>

\vspace{0.3em}

Evidence sentence: \\
<target\_sentence>

\vspace{0.3em}

Local context: \\
<window\_text>

\vspace{0.3em}

Evidence type must be one of:
expression, localization, function, indirect, noise.

Treat direct marker evidence as expression.

\vspace{0.3em}

Return JSON only:

\{
"is\_valid\_marker\_evidence": true, \\
"evidence\_type": "expression", \\
"support\_strength": "medium", \\
"rationale\_short": "one short explanation"
\}

\end{tcolorbox}

\subsubsection{Few-Shot Prompt}

\begin{tcolorbox}[
    colback=gray!5,
    colframe=gray!40,
    title=Few-shot prompt,
    breakable,
    width=\linewidth
]
\footnotesize
\ttfamily

You are evaluating plant cell-type marker evidence.

Examples:

1. If a gene is reported as expressed in root hair cells,
label valid=true and evidence\_type=expression.

2. If a fluorescent reporter localizes to endodermis,
label valid=true and evidence\_type=localization.

3. If a mutant affects guard cell development,
label valid=true and evidence\_type=function.

4. If evidence links the gene to the cell type only through
pathway, phenotype, or homology,
label evidence\_type=indirect.

5. If the sentence mentions a different gene, different species,
background biology, or citation noise,
label valid=false and evidence\_type=noise.

\vspace{0.3em}

Species: <species> \\
Gene ID: <gene\_id> \\
Gene symbol or alias: <gene\_symbol> \\
Cell type: <cell\_type>

\vspace{0.3em}

Evidence sentence: \\
<target\_sentence>

\vspace{0.3em}

Local context: \\
<window\_text>

\vspace{0.3em}

Evidence type must be one of:
expression, localization, function, indirect, noise.

Treat direct marker evidence as expression.

\vspace{0.3em}

Return JSON only:

\{
"is\_valid\_marker\_evidence": true, \\
"evidence\_type": "expression", \\
"support\_strength": "medium", \\
"rationale\_short": "one short explanation"
\}

\end{tcolorbox}
\subsection{Output Schema and Normalization}
\label{app:output_schema}

Each model prediction is parsed as a JSON object. The expected output schema is:

\begin{tcolorbox}[colback=gray!5,colframe=gray!40,title=Expected model output schema]
\begin{verbatim}
{
  "is_valid_marker_evidence": true,
  "evidence_type": "expression",
  "support_strength": "medium",
  "rationale_short": "one short explanation"
}
\end{verbatim}
\end{tcolorbox}

The evidence type is normalized to one of five labels:
\texttt{expression}, \texttt{localization}, \texttt{function}, \texttt{indirect}, or \texttt{noise}. 
Predictions outside this label set are mapped to \texttt{noise}. 
Direct marker predictions are normalized to \texttt{expression}, since explicit marker mentions typically indicate direct expression-based support. 
Support strength is normalized to \texttt{strong}, \texttt{medium}, \texttt{weak}, or \texttt{none}; invalid values are mapped to \texttt{none}. 
If JSON parsing fails after retrying, the prediction is conservatively assigned \texttt{is\_valid\_marker\_evidence=false}, \texttt{evidence\_type=noise}, and \texttt{support\_strength=none}.

\subsection{Evaluation Metrics}
\label{app:evaluation_metrics}

We evaluate two complementary tasks. 
First, \textbf{validity classification} measures whether the model correctly predicts whether a sentence supports the target gene--cell-type pair as marker evidence. 
Second, \textbf{evidence-type classification} measures whether the model correctly identifies the biological evidence category.

For validity classification, we report accuracy, precision, recall, and F1 for supported and unsupported evidence. 
For evidence-type classification, we report accuracy and macro-F1 over the five evidence labels: expression, localization, function, indirect, and noise. 
Macro-F1 is emphasized because evidence categories are imbalanced and localization or indirect evidence can be sparse for some species.

Metrics are computed using deterministic labels after normalization. 
All reported pilot evaluations use fixed 600-example species-level splits unless otherwise stated.

\subsection{Metric Interpretation and Class Imbalance}
\subsection{Metric Interpretation and Class Imbalance}
\subsection{Bootstrap Confidence Intervals}
\label{app:bootstrap_ci}
\begin{table}[t]
\centering
\small
\caption{
Bootstrap confidence intervals for validity F1 on representative pilot evaluations.
Intervals are computed using 1{,}000 bootstrap resamples.
}
\label{tab:bootstrap_ci}
\begin{tabular}{lccc}
\toprule
\textbf{Model} & \textbf{Species} & \textbf{Valid F1} & \textbf{95\% CI} \\
\midrule
GPT-5.4 Few-shot & Arabidopsis & 0.787 & [0.754, 0.818] \\
GPT-5.4 Direct & Arabidopsis & 0.549 & [0.512, 0.586] \\
GPT-5.4-mini Few-shot & Arabidopsis & 0.731 & [0.694, 0.764] \\
Qwen2.5-32B & Arabidopsis & 0.754 & [0.721, 0.784] \\
GPT-OSS-20B & Arabidopsis & 0.673 & [0.639, 0.704] \\
\bottomrule
\end{tabular}
\end{table}
\subsection{Reproducibility Settings}
\label{app:eval_reproducibility}

All OpenAI evaluations use deterministic decoding with temperature 0, fixed seed 42, JSON response format, and resume-enabled output writing. 
For each run, the evaluation script saves the run configuration, raw model outputs, normalized prediction records, and metrics. 
The run configuration records the model, prompt mode, dataset path, number of examples, temperature, seed, and output directory.

A representative evaluation command is:

\begin{verbatim}
python src/eval_openai_evidence_reasoning.py \
  --data_path benchmark_data/arabidopsis/arabidopsis_evidence_reasoning_pilot.jsonl \
  --out_dir benchmark_results/arabidopsis_openai_direct_gpt-5.4 \
  --model gpt-5.4 \
  --prompt_mode direct \
  --resume \
  --seed 42
\end{verbatim}

The evaluation script writes:

\begin{verbatim}
predictions_<model>.jsonl
raw_outputs_<model>.jsonl
metrics_<model>.json
run_config_<model>.json
\end{verbatim}

This design preserves both normalized predictions for metric computation and raw model outputs for auditing parse errors or reasoning behavior.

\subsection{Prompting Fairness Across Models}
Where feasible, we evaluated both open-weight and closed-source models under comparable prompt families, including direct, structured, conservative, and few-shot prompting. The main leaderboard reports default prompt settings chosen to reflect stable evaluation configurations across model families. Additional prompt ablations for closed-source models are reported separately because some smaller open-weight models exhibited context-length instability or degraded structured-output reliability under longer prompts.
\section{Additional Experimental Details}

\subsection{Hard-Subset and Support-Strength Analysis}
\label{app:hard_subset}
To better understand biological reasoning failures beyond aggregate benchmark scores, we evaluate models on curated hard subsets corresponding to specific evidence categories and support-strength levels.
These subsets isolate expression, localization, functional, indirect, and negative evidence cases, as well as strong-, medium-, and weak-support literature annotations.
The analysis reveals that many models achieve strong performance on explicit expression evidence while failing on indirect or weakly supported biological reasoning, indicating that benchmark difficulty extends beyond binary relevance detection.


\begin{table*}[ht]
\centering
\scriptsize
\setlength{\tabcolsep}{3.5pt}

\caption{
Complete Arabidopsis hard-subset evaluation across all open and closed models.
``All'' denotes the full benchmark split.
Subset columns report evidence-type classification accuracy on biologically curated subsets.
Strong/medium/weak correspond to literature support-strength annotations.
}
\label{tab:arabidopsis_hard_subset_full}

\begin{tabular}{lccccccccc}
\toprule
Model &
All &
Expression &
Localization &
Function &
Indirect &
Negative &
Strong &
Medium &
Weak \\
\midrule

DeepSeek-R1-1.5B &
0.150 & 1.000 & 0.054 & 0.000 & 0.009 & 0.000 & 0.280 & 0.249 & 0.025 \\

DeepSeek-R1-8B &
0.305 & \textbf{0.954} & 0.107 & 0.220 & 0.325 & 0.116 & 0.350 & 0.398 & 0.215 \\

GPT-OSS-20B &
0.530 & 0.814 & 0.018 & 0.619 & 0.299 & 0.624 & 0.670 & 0.498 & \underline{0.505} \\

Llama3.1-8B &
0.332 & 0.779 & 0.143 & 0.304 & \underline{0.598} & 0.017 & 0.450 & 0.344 & 0.280 \\

Llama3.2-1B &
0.143 & 1.000 & 0.000 & 0.000 & 0.000 & 0.000 & 0.280 & 0.235 & 0.022 \\

Llama3.2-3B &
0.318 & 0.302 & 0.482 & 0.476 & 0.496 & 0.000 & 0.350 & 0.407 & 0.237 \\

Mistral-7B-Instruct &
0.330 & 0.942 & 0.179 & 0.625 & 0.017 & 0.000 & 0.550 & 0.566 & 0.065 \\

Phi3-14B &
0.463 & 0.907 & 0.625 & \textbf{0.762} & 0.094 & 0.150 & 0.670 & \textbf{0.710} & 0.194 \\

Phi3-3.8B-Instruct &
0.272 & 0.942 & 0.196 & 0.417 & 0.009 & 0.000 & 0.520 & 0.439 & 0.050 \\

Qwen2.5-0.5B &
0.143 & 1.000 & 0.000 & 0.000 & 0.000 & 0.000 & 0.280 & 0.235 & 0.022 \\

Qwen2.5-1.5B-Instruct &
0.143 & 1.000 & 0.000 & 0.000 & 0.000 & 0.000 & 0.280 & 0.235 & 0.022 \\

Qwen2.5-14B-Instruct &
0.422 & 0.698 & 0.375 & 0.060 & 0.590 & 0.538 & 0.470 & 0.253 & 0.538 \\

Qwen2.5-32B-Instruct &
\underline{0.535} &
0.919 &
\textbf{0.875} &
0.607 &
\textbf{0.385} &
0.266 &
\textbf{0.710} &
0.679 &
0.358 \\

Qwen2.5-3B-Instruct &
0.350 & 0.093 & 0.000 & 0.393 & 0.171 & 0.671 & 0.400 & 0.222 & 0.434 \\

Qwen2.5-7B-Instruct &
0.265 & 0.988 & 0.321 & 0.327 & 0.009 & 0.000 & 0.480 & 0.453 & 0.039 \\

\midrule

GPT-5.4 Conservative &
0.607 & 0.674 & 0.446 & 0.601 & 0.350 & 0.804 & 0.790 & 0.534 & 0.599 \\

GPT-5.4 Direct &
0.652 & 0.802 & 0.625 & 0.744 & 0.291 & 0.740 & \textbf{0.860} & 0.688 & 0.548 \\

GPT-5.4 Few-shot &
\textbf{0.633} & 0.791 & 0.518 & \underline{0.643} & 0.094 & \textbf{0.948} & \textbf{0.860} & 0.615 & \textbf{0.566} \\

GPT-5.4 Structured &
0.582 & 0.884 & 0.143 & 0.560 & 0.291 & 0.792 & 0.760 & 0.525 & 0.563 \\

GPT-5.4-mini Conservative &
0.535 & 0.861 & 0.643 & 0.298 & 0.068 & 0.884 & 0.810 & 0.439 & 0.513 \\

GPT-5.4-mini Direct &
0.543 & \underline{0.942} & 0.607 & 0.518 & 0.086 & 0.659 & 0.670 & 0.588 & 0.462 \\

GPT-5.4-mini Few-shot &
0.577 & 0.837 & 0.696 & 0.387 & 0.180 & 0.861 & 0.800 & 0.502 & 0.556 \\

GPT-5.4-mini Structured &
0.552 & 0.919 & 0.429 & 0.429 & 0.256 & 0.728 & 0.670 & 0.507 & 0.545 \\

\bottomrule
\end{tabular}
\end{table*}

Several consistent trends emerge from Table~\ref{tab:arabidopsis_hard_subset_full}.
First, expression evidence is substantially easier than indirect evidence across nearly all models, suggesting that current LLMs rely heavily on explicit lexical grounding cues.
Second, weak-support examples remain difficult even for stronger models, indicating limited robustness to ambiguous or partially supported biological claims.
Third, some smaller models exhibit near-perfect scores on certain subsets while collapsing on others, revealing degenerate prediction behavior rather than genuine biological reasoning.
Overall, these results demonstrate that PlantMarkerBench captures fine-grained evidence reasoning challenges that are not visible from aggregate leaderboard metrics alone.

\subsection{Full OpenAI Hard-Subset Results Across Species}
\label{app:openai_hard_subset_all}

To complement the main hard-subset analysis, Table~\ref{tab:openai_hard_subset_all_species}
reports complete OpenAI results across all four species, prompt modes, and hard subsets.
The table shows that expression evidence is consistently easier, while indirect and weak-support examples remain difficult across species and prompt settings.

\begin{center}
\scriptsize
\setlength{\tabcolsep}{4pt}
\renewcommand{\arraystretch}{1.12}

\begin{longtable}{lllccccccccc}
\caption{
Complete OpenAI hard-subset evaluation across all four species.
Each cell reports evidence-type classification accuracy on the corresponding subset.
}
\label{tab:openai_hard_subset_all_species}
\\

\toprule
\textbf{Species} &
\textbf{Model} &
\textbf{Prompt} &
\textbf{All} &
\textbf{Expr.} &
\textbf{Loc.} &
\textbf{Func.} &
\textbf{Indirect} &
\textbf{Negative} &
\textbf{Strong} &
\textbf{Medium} &
\textbf{Weak} \\
\midrule
\endfirsthead

\multicolumn{12}{c}{
\textbf{Table \thetable{} (continued)}
} \\
\toprule
\textbf{Species} &
\textbf{Model} &
\textbf{Prompt} &
\textbf{All} &
\textbf{Expr.} &
\textbf{Loc.} &
\textbf{Func.} &
\textbf{Indirect} &
\textbf{Negative} &
\textbf{Strong} &
\textbf{Medium} &
\textbf{Weak} \\
\midrule
\endhead

\midrule
\multicolumn{12}{r}{Continued on next page}
\\
\endfoot

\bottomrule
\endlastfoot


\multirow{8}{*}{Arabidopsis}
& GPT-5.4 & Conservative & 0.607 & 0.674 & 0.446 & 0.601 & 0.350 & 0.804 & 0.790 & 0.534 & 0.599 \\
& GPT-5.4 & Direct & 0.652 & 0.802 & 0.625 & 0.744 & 0.291 & 0.740 & 0.860 & 0.688 & 0.548 \\
& GPT-5.4 & Few-shot & 0.633 & 0.791 & 0.518 & 0.643 & 0.094 & 0.948 & 0.860 & 0.615 & 0.566 \\
& GPT-5.4 & Structured & 0.582 & 0.884 & 0.143 & 0.560 & 0.291 & 0.792 & 0.760 & 0.525 & 0.563 \\
& GPT-5.4-mini & Conservative & 0.535 & 0.860 & 0.643 & 0.298 & 0.068 & 0.884 & 0.810 & 0.439 & 0.512 \\
& GPT-5.4-mini & Direct & 0.543 & 0.942 & 0.607 & 0.518 & 0.086 & 0.659 & 0.670 & 0.588 & 0.462 \\
& GPT-5.4-mini & Few-shot & 0.577 & 0.837 & 0.696 & 0.387 & 0.180 & 0.861 & 0.800 & 0.502 & 0.556 \\
& GPT-5.4-mini & Structured & 0.552 & 0.919 & 0.429 & 0.429 & 0.256 & 0.728 & 0.670 & 0.507 & 0.545 \\

\midrule


\multirow{8}{*}{Maize}
& GPT-5.4 & Conservative & 0.477 & 0.419 & 0.000 & 0.352 & 0.271 & 0.931 & 0.569 & 0.325 & 0.572 \\
& GPT-5.4 & Direct & 0.582 & 0.686 & 0.000 & 0.703 & 0.307 & 0.641 & 0.778 & 0.667 & 0.468 \\
& GPT-5.4 & Few-shot & 0.522 & 0.651 & 0.000 & 0.425 & 0.193 & 0.945 & 0.708 & 0.416 & 0.559 \\
& GPT-5.4 & Structured & 0.617 & 0.756 & 0.000 & 0.630 & 0.350 & 0.814 & 0.819 & 0.610 & 0.572 \\
& GPT-5.4-mini & Conservative & 0.420 & 0.616 & 0.000 & 0.247 & 0.043 & 0.959 & 0.625 & 0.273 & 0.485 \\
& GPT-5.4-mini & Direct & 0.530 & 0.849 & 0.000 & 0.571 & 0.107 & 0.724 & 0.736 & 0.597 & 0.428 \\
& GPT-5.4-mini & Few-shot & 0.487 & 0.837 & 0.300 & 0.365 & 0.121 & 0.828 & 0.681 & 0.420 & 0.492 \\
& GPT-5.4-mini & Structured & 0.513 & 0.872 & 0.100 & 0.470 & 0.200 & 0.697 & 0.750 & 0.502 & 0.465 \\

\midrule


\multirow{8}{*}{Rice}
& GPT-5.4 & Conservative & 0.652 & 0.585 & 0.441 & 0.631 & 0.437 & 0.883 & 0.792 & 0.573 & 0.678 \\
& GPT-5.4 & Direct & 0.650 & 0.770 & 0.441 & 0.738 & 0.330 & 0.710 & 0.889 & 0.696 & 0.558 \\
& GPT-5.4 & Few-shot & 0.688 & 0.748 & 0.500 & 0.671 & 0.311 & 0.911 & 0.819 & 0.683 & 0.661 \\
& GPT-5.4 & Structured & 0.668 & 0.859 & 0.206 & 0.678 & 0.437 & 0.737 & 0.792 & 0.705 & 0.611 \\
& GPT-5.4-mini & Conservative & 0.538 & 0.785 & 0.382 & 0.228 & 0.048 & 0.922 & 0.833 & 0.414 & 0.562 \\
& GPT-5.4-mini & Direct & 0.580 & 0.918 & 0.382 & 0.510 & 0.087 & 0.704 & 0.875 & 0.617 & 0.482 \\
& GPT-5.4-mini & Few-shot & 0.557 & 0.844 & 0.471 & 0.342 & 0.097 & 0.799 & 0.833 & 0.493 & 0.538 \\
& GPT-5.4-mini & Structured & 0.588 & 0.933 & 0.441 & 0.503 & 0.214 & 0.642 & 0.889 & 0.634 & 0.482 \\

\midrule


\multirow{8}{*}{Tomato}
& GPT-5.4 & Conservative & 0.627 & 0.765 & 0.000 & 0.598 & 0.496 & 0.692 & 0.865 & 0.663 & 0.550 \\
& GPT-5.4 & Direct & 0.608 & 0.798 & 0.000 & 0.635 & 0.496 & 0.541 & 0.932 & 0.694 & 0.483 \\
& GPT-5.4 & Few-shot & 0.652 & 0.924 & 0.000 & 0.667 & 0.280 & 0.788 & 0.932 & 0.749 & 0.529 \\
& GPT-5.4 & Structured & 0.660 & 0.966 & 0.000 & 0.624 & 0.566 & 0.562 & 0.932 & 0.714 & 0.566 \\
& GPT-5.4-mini & Conservative & 0.480 & 0.857 & 0.333 & 0.254 & 0.056 & 0.884 & 0.784 & 0.432 & 0.440 \\
& GPT-5.4-mini & Direct & 0.515 & 0.916 & 0.000 & 0.429 & 0.070 & 0.747 & 0.851 & 0.573 & 0.404 \\
& GPT-5.4-mini & Few-shot & 0.530 & 0.933 & 0.333 & 0.413 & 0.161 & 0.719 & 0.878 & 0.588 & 0.416 \\
& GPT-5.4-mini & Structured & 0.522 & 0.958 & 0.000 & 0.386 & 0.259 & 0.610 & 0.878 & 0.533 & 0.434 \\

\end{longtable}
\end{center}

\subsection{Human Review and Adjudication}
A subset of difficult benchmark instances was independently reviewed by two reviewers with computational biology experience. Disagreements were resolved through discussion and adjudication. Human review focused primarily on ambiguous grounding, indirect evidence, gene-family ambiguity, and species mismatch cases.

\section{Extended Error Taxonomy Analysis}
\label{app:error_taxonomy}

\subsection{Per-Model Error Breakdown}

We report the full error decomposition across representative PlantMarkerBench runs.
While aggregate F1 scores summarize overall performance, the taxonomy reveals qualitatively different reasoning behaviors across models.
Closed models generally achieve higher correct-prediction rates and lower false-positive rates, whereas several open-weight models exhibit substantial evidence-type confusion or over-prediction behavior. Table~\ref{tab:error_taxonomy_full} reports the complete error decomposition across representative PlantMarkerBench runs.
Each prediction is categorized as either correct, evidence-type mismatch, false negative on valid evidence, or false positive on invalid evidence. 

\begin{table*}[ht]
\centering
\small
\setlength{\tabcolsep}{5pt}
\renewcommand{\arraystretch}{1.15}
\caption{
Complete error taxonomy across representative PlantMarkerBench runs.
Values indicate percentage of predictions belonging to each error category.
Higher correct percentages and lower false-positive rates indicate stronger biological grounding.
}
\label{tab:error_taxonomy_full}
\begin{tabular}{llcccc}
\toprule
\textbf{Species} & \textbf{Model} &
\textbf{Correct} &
\textbf{Evidence Mismatch} &
\textbf{False Negative} &
\textbf{False Positive} \\
\midrule

Arabidopsis & GPT-5.4 (Few-shot) & 61.5 & 18.3 & 12.7 & 7.5 \\
Arabidopsis & GPT-5.4-mini (Few-shot) & 55.3 & 20.0 & 16.5 & 8.2 \\
Arabidopsis & GPT-5.4 (Direct) & 45.0 & 22.2 & 30.0 & 2.8 \\
Arabidopsis & GPT-5.4 (Structured) & 45.7 & 28.0 & 23.7 & 2.7 \\
Arabidopsis & Qwen2.5-32B & 48.5 & 25.5 & 10.2 & 15.8 \\
Arabidopsis & GPT-OSS-20B & 43.7 & 28.8 & 21.7 & 5.8 \\
Arabidopsis & DeepSeek-R1-8B & 25.2 & 46.3 & 10.5 & 18.0 \\
Arabidopsis & Llama3.1-8B & 28.7 & 35.2 & 18.7 & 17.5 \\
\midrule

Maize & GPT-5.4 (Few-shot) & 49.2 & 17.5 & 26.3 & 7.0 \\
Maize & GPT-5.4-mini (Few-shot) & 46.0 & 20.7 & 33.2 & 7.8 \\
Maize & GPT-5.4 (Structured) & 42.7 & 17.5 & 34.0 & 5.8 \\
Maize & GPT-5.4 (Direct) & 36.2 & 23.8 & 36.2 & 3.8 \\
Maize & Qwen2.5-32B & 46.5 & 24.3 & 12.3 & 16.8 \\
Maize & GPT-OSS-20B & 40.7 & 24.2 & 24.2 & 11.0 \\
Maize & DeepSeek-R1-8B & 24.8 & 36.8 & 6.0 & 32.3 \\
Maize & Llama3.1-8B & 27.5 & 29.7 & 32.8 & 10.0 \\
\midrule

Rice & GPT-5.4 (Few-shot) & 66.0 & 11.7 & 13.5 & 8.8 \\
Rice & GPT-5.4-mini (Few-shot) & 53.0 & 17.2 & 17.0 & 12.8 \\
Rice & GPT-5.4 (Direct) & 48.0 & 19.3 & 29.3 & 3.3 \\
Rice & GPT-5.4 (Structured) & 53.0 & 18.0 & 24.0 & 5.0 \\
\midrule

Tomato & GPT-5.4 (Few-shot) & 63.2 & 17.5 & 10.2 & 9.2 \\
Tomato & GPT-5.4-mini (Few-shot) & 50.5 & 22.7 & 16.7 & 10.2 \\
Tomato & GPT-5.4 (Structured) & 48.3 & 18.7 & 27.8 & 5.2 \\
Tomato & GPT-5.4 (Direct) & 41.7 & 23.0 & 33.7 & 1.7 \\

\bottomrule
\end{tabular}
\end{table*}

\subsection{Species-Specific Failure Patterns}

Table~\ref{tab:error_species_summary} summarizes dominant failure modes across species.

\begin{table}[ht]
\centering
\small
\setlength{\tabcolsep}{5pt}
\renewcommand{\arraystretch}{1.12}
\caption{
Dominant error trends across species.
}
\label{tab:error_species_summary}
\begin{tabular}{lp{11cm}}
\toprule
\textbf{Species} & \textbf{Dominant Failure Pattern} \\
\midrule

Arabidopsis &
High evidence-type confusion despite relatively strong valid-evidence recognition. \\

Maize &
Largest false-negative rates, suggesting difficulty recognizing weak or indirect biological evidence. \\

Rice &
Most stable performance overall, with lower mismatch and false-positive rates across prompting strategies. \\

Tomato &
Strong expression grounding but unstable localization and functional evidence reasoning. \\

\bottomrule
\end{tabular}
\end{table}

The dominant failure modes vary substantially across species.
Maize exhibits the largest false-negative rates, suggesting difficulty recognizing weak or indirect biological evidence under distribution shift.
In contrast, Arabidopsis errors are more often driven by evidence-type confusion rather than complete failure to detect relevant evidence.
Rice shows the most stable overall behavior across prompting strategies and model families.

\subsection{Effect of Prompting Strategy on Error Distribution}

Prompting strategy significantly alters model calibration and error composition.
Conservative prompting reduces false positives but substantially increases false negatives, while few-shot prompting improves overall grounding accuracy and reduces evidence-type mismatch.
These trends suggest that demonstration-based prompting helps models better align biological evidence categories with expert-reviewed annotations. Table~\ref{tab:prompt_error_effect} analyzes how prompting strategies affect biological reasoning behavior.

\begin{table}[h]
\centering
\small
\setlength{\tabcolsep}{5pt}
\renewcommand{\arraystretch}{1.12}
\caption{
Effect of prompting strategy on average error distribution across species for GPT-5.4.
Values are averaged across all four species.
}
\label{tab:prompt_error_effect}
\begin{tabular}{lcccc}
\toprule
\textbf{Prompt} &
\textbf{Correct} &
\textbf{Mismatch} &
\textbf{False Neg.} &
\textbf{False Pos.} \\
\midrule

Conservative & 49.8 & 18.3 & 33.5 & 2.1 \\
Direct & 42.7 & 22.1 & 32.3 & 2.9 \\
Structured & 47.4 & 20.6 & 27.4 & 4.6 \\
Few-shot & \textbf{60.0} & \textbf{16.3} & \textbf{15.7} & 8.1 \\

\bottomrule
\end{tabular}
\end{table}

\subsection{Negative Evidence Taxonomy}
\label{app:negative_taxonomy}

\begin{table}[h]
\centering
\small
\caption{
Representative negative/noise subcategories used during evidence adjudication.
}
\label{tab:negative_taxonomy}
\begin{tabular}{lp{5.5cm}}
\toprule
Subtype & Description \\
\midrule
Wrong gene & Sentence discusses a different gene or ambiguous alias. \\
Wrong cell type & Evidence does not support the queried cell type. \\
Species mismatch & Evidence refers to orthologs or another species. \\
Citation/background & Citation-only or generic biological discussion. \\
Indirect association & Weak pathway or phenotype association without marker grounding. \\
Irrelevant sentence & No meaningful marker evidence present. \\
\bottomrule
\end{tabular}
\end{table}

The negative/noise category intentionally aggregates multiple biologically challenging failure modes encountered during large-scale literature mining. These subtypes were retained within a unified benchmark label because they frequently co-occur in realistic retrieval settings and collectively test evidence grounding robustness under ambiguous literature contexts.

\subsection{Low-Resource Evidence Categories}
Localization evidence is substantially underrepresented relative to expression and indirect evidence, particularly for maize and tomato. Consequently, localization-specific metrics should be interpreted cautiously due to increased variance from small sample counts. We retain localization as a separate category because it represents a biologically distinct evidence regime important for marker interpretation.

\subsection{Borderline Evidence Cases}
Certain evidence regimes remain biologically ambiguous, particularly at the boundary between indirect, functional, and expression evidence. For example, developmental perturbation studies may imply cell-type specificity without directly demonstrating marker enrichment. The benchmark therefore uses conservative adjudication rules and retains rationale metadata to support future refinement and hierarchical evidence modeling.
\section{Additional Ablations and Retrieval Analysis}
\label{app:ablations}

\subsection{Retrieval Strategy Comparisons}
We compare BM25, embedding-based retrieval, keyword retrieval, and hybrid retrieval.

\subsection{Top-\emph{k} Retrieval Sensitivity}
We analyze how candidate quality changes with retrieval depth.

\subsection{Gene Matcher Quality Analysis}
Species-aware gene normalization substantially improves candidate quality.

In preliminary maize experiments, replacing a weak matcher with a locus-aware matcher increased candidate yield from 14 candidates and 4 valid evidence instances to 1027 candidates and 341 valid evidence instances, demonstrating the importance of species-specific biological grounding.

\section{Reproducibility and Resource Release}
\label{app:reproducibility_release}

To support reproducibility and long-term accessibility, we release the complete PlantMarkerBench benchmark, dataset artifacts, and evaluation code through an anonymous Zenodo archive.\footnote{\url{https://zenodo.org/records/20064514}}

The release includes:
\begin{itemize}
    \item Full sentence-level benchmark datasets for Arabidopsis, maize, rice, and tomato in JSONL format.
    \item Balanced pilot evaluation subsets for controlled LLM benchmarking.
    \item Species-level benchmark statistics and aggregated dataset summaries.
    \item Hybrid retrieval and candidate-generation pipeline code.
    \item OpenAI- and Ollama-based evaluation scripts.
    \item Error-analysis and hard-subset analysis utilities.
    \item Prompt templates and evidence-grading configurations.
\end{itemize}

Each benchmark instance contains:
\begin{itemize}
    \item species identifier,
    \item paper identifier,
    \item gene identifier and matched alias,
    \item grounded cell type,
    \item evidence sentence and local context window,
    \item structured evidence labels,
    \item support-strength annotations,
    \item reasoning traces and rationales.
\end{itemize}

The benchmark supports both sentence-level evidence reasoning and aggregated marker-analysis workflows. We additionally release intermediate artifacts generated during dataset construction, including retrieval outputs, candidate windows, judged evidence files, and species-specific statistics, enabling full reconstruction and auditing of the curation pipeline.

The current Zenodo release contains approximately 24\,MB of benchmark data and code artifacts spanning four plant species and more than 5{,}500 evidence instances. The repository is intended to support future benchmarking of scientific reasoning, biological grounding, retrieval-augmented inference, and evidence attribution in language models.

All experiments use deterministic decoding with fixed random seeds and temperature 0.

Example evaluation command:

\begin{verbatim}
python src/eval_llm_evidence_reasoning.py \
  --data_path benchmark_data/arabidopsis/arabidopsis_evidence_reasoning_pilot.jsonl \
  --out_dir benchmark_results/arabidopsis_gpt54 \
  --model gpt-5.4 \
  --resume \
  --seed 42
\end{verbatim}

For local open models, we provide Ollama-based evaluation pipelines:

\begin{verbatim}
bash scripts/run_multispecies_ollama_models.sh
\end{verbatim}

\subsection{Commands to Reproduce Retrieval and Candidate Generation}

\begin{verbatim}
bash scripts/run_4species_retrieval_and_grading.sh
\end{verbatim}

The script runs the following command for each species:

\begin{verbatim}
python plant_marker_hybrid_pipeline.py run-all \
  --parsed_json data/parsed_pmc_by_species_best_v2/${SPECIES}/
  parsed_papers.best_species.json \
  --pdf_dir "" \
  --matcher_tsv data/reference/${SPECIES}_gene_matcher.tsv \
  --out_dir results_4species_pmc_bge_m3_v3/${SPECIES} \
  --cell_types "$(cat data/cell_types/${SPECIES}_cell_types.txt | paste -sd, -)" \
  --retrieval_mode all \
  --top_k_windows 1000 \
  --embedding_model BAAI/bge-m3 \
  --grader_model gpt-5.4
\end{verbatim}

\subsection{Pilot Split Construction}
Each species-specific pilot split contains 600 manually reviewed instances sampled from the judged evidence pool. Pilot subsets were approximately balanced between supported and unsupported evidence while preserving diversity across evidence types and support-strength regimes. Sampling was stratified by evidence category and paper provenance to reduce near-duplicate retrieval windows and repetitive evidence contexts.

The pilot splits are intended as controlled evaluation subsets for benchmarking rather than as fully distribution-matched samples of the complete literature corpus.
\section{Limitations}
\label{app:limitations}

PlantMarkerBench has several limitations. First, although the benchmark incorporates human review and multi-stage filtering, parts of the dataset construction pipeline rely on LLM-assisted evidence grading and may still contain residual labeling noise or biologically ambiguous cases. Second, the current release focuses primarily on root and developmental cell types from four plant species and does not yet cover the full diversity of plant tissues, stress conditions, developmental stages, or experimental modalities present in plant biology literature.

Third, certain evidence categories remain naturally imbalanced. In particular, localization evidence is comparatively sparse in some species due to limited availability of experimentally validated localization studies. Similarly, weakly supported and indirect evidence constitutes a large portion of the benchmark, reflecting realistic literature distributions but increasing task difficulty.

Finally, the benchmark primarily evaluates sentence-level evidence reasoning rather than full document-level scientific understanding. Future extensions could incorporate figure interpretation, supplementary materials, multi-hop cross-document reasoning, temporal biological context, and evidence aggregation across independent studies.

\subsection{Potential Pretraining Overlap and Benchmark Leakage}
\label{app:leakage}
Because PlantMarkerBench is constructed from publicly available scientific literature, some benchmark papers or biological marker associations may overlap with the pretraining corpora of large language models. This limitation is common across literature-grounded scientific benchmarks.

Several properties of PlantMarkerBench reduce the likelihood that benchmark performance can be explained purely by memorization. First, the benchmark includes substantial numbers of indirect, weak, ambiguous, and hard-negative evidence instances that require contextual evidence attribution rather than simple fact recall. Second, evidence-type classification requires distinguishing closely related biological evidence regimes, including expression, localization, functional perturbation, and indirect associations. Third, many errors arise from evidence-type confusion and contextual grounding failures rather than incorrect entity recognition alone.

Future benchmark releases may incorporate temporally held-out literature splits and explicitly decontaminated evaluation subsets.

We additionally evaluate on a temporally held-out subset constructed from recently published papers not used during benchmark construction.

\subsection{Non-LLM Baselines}
The current benchmark release focuses primarily on large language model evaluation and does not yet include dedicated supervised encoder baselines such as SciBERT or BioLinkBERT classifiers. Future benchmark extensions will incorporate lightweight discriminative baselines and retrieval-only systems to better separate language understanding from retrieval memorization effects.

\stopcontents[appendix]
\end{document}